\documentclass[sigconf,authorversion]{acmart}
\AtBeginDocument{%
  }

\usepackage{multirow}
\usepackage{amsmath}
\usepackage{cleveref}
\usepackage{xcolor}
\usepackage{framed}
\definecolor{shadecolor}{gray}{0.9} 
\usepackage{graphicx}

\copyrightyear{2026}
\acmYear{2026}
\acmNumpages{12}
\setcopyright{cc}
\acmConference[VRST '26]{32nd ACM Symposium on Virtual Reality Software and Technology}{November 16--18, 2026}{Sendai, Japan}
\acmBooktitle{32nd ACM Symposium on Virtual Reality Software and Technology (VRST '26), November 16--18, 2026, Sendai, Japan}
\acmDOI{10.1145/3822517.3848685}
\acmISBN{979-8-4007-2811-2/2026/11}

\acmSubmissionID{1208}

\begin{document}

\title[VLMs for Perceptual Quality Assessment and Content Adjustment in AR]{Harnessing Vision-Language Models for Perceptual Quality Assessment and Autonomous Content Adjustment in Augmented Reality}

\author{Elias Rotondo}
\authornote{Equal contribution.}
\orcid{0000-0002-9674-6135}
\affiliation{
  \institution{Duke University}
  \city{Durham}
  \state{NC}
  \country{USA}
}
\email{eli.rotondo@duke.edu}

\author{Lin Duan}
\authornotemark[1]
\orcid{0000-0001-8515-6351}
\affiliation{
  \institution{Duke University}
  \city{Durham}
  \state{NC}
  \country{USA}
}
\email{lin.duan@duke.edu}

\author{Yanming Xiu}
\orcid{0009-0008-1547-3261}
\affiliation{
  \institution{Duke University}
  \city{Durham}
  \state{NC}
  \country{USA}
}
\email{yanming.xiu@duke.edu}

\author{Sangjun Eom}
\orcid{0000-0003-2075-6932}
\affiliation{
  \institution{Duke University}
  \city{Durham}
  \state{NC}
  \country{USA}
}
\email{sangjun.eom@duke.edu}

\author{Conrad Li}
\orcid{0009-0004-6436-8258}
\affiliation{
  \institution{Duke University}
  \city{Durham}
  \state{NC}
  \country{USA}
}
\email{conrad.li@duke.edu}

\author{Maria Gorlatova}
\orcid{0000-0002-5477-7830}
\affiliation{
  \institution{Duke University}
  \city{Durham}
  \state{NC}
  \country{USA}
}
\email{maria.gorlatova@duke.edu}

\renewcommand{\shortauthors}{Rotondo et al.}

\begin{abstract}
  Advancements in augmented reality (AR) continue to foster innovative solutions, facilitating novel methodologies within educational systems, healthcare delivery, and risk-mitigation protocols. However, optimizing for end-user immersion and comfort remains challenging, as AR head-mounted displays contend with constrained scene geometry, spatial jitter, and temporal instability. User studies are the standard AR evaluation method for visual quality, but their cost, diminishing scalability, and inflexibility pose bottlenecks during iterative application design. To address this problem, we present an automated framework for AR content evaluation and refinement, built on vision-language models (VLMs), to evaluate and predict the visual fidelity of AR scenes as perceived by users. First, we introduce RateAR, a benchmark of AR images and videos collected across diverse scenes and environmental conditions, with good-to-excellent reliability ($ICC(2,5)\geq.90$) across perceptual factors, including object placement, scale, and shadow consistency. Subsequently, we evaluate eleven commercial VLMs on the crafted benchmark. Results support that VLM-based quality predictions strongly correlate with human subjective judgments, achieving Spearman’s rank-order correlations of up to 0.8695. An ablation study further suggests that, compared to other prompting strategies, our contextual prompting yields better alignment with human ratings while balancing introduced complexity cues. Building on these findings, we construct an automated AR content adjustment system and conduct a 21-participant user study. More than 90\% of participants found that the system improved placement and size coherence of virtual content.
\end{abstract}



\begin{teaserfigure}
 \centering
 \includegraphics[width=0.90\textwidth]{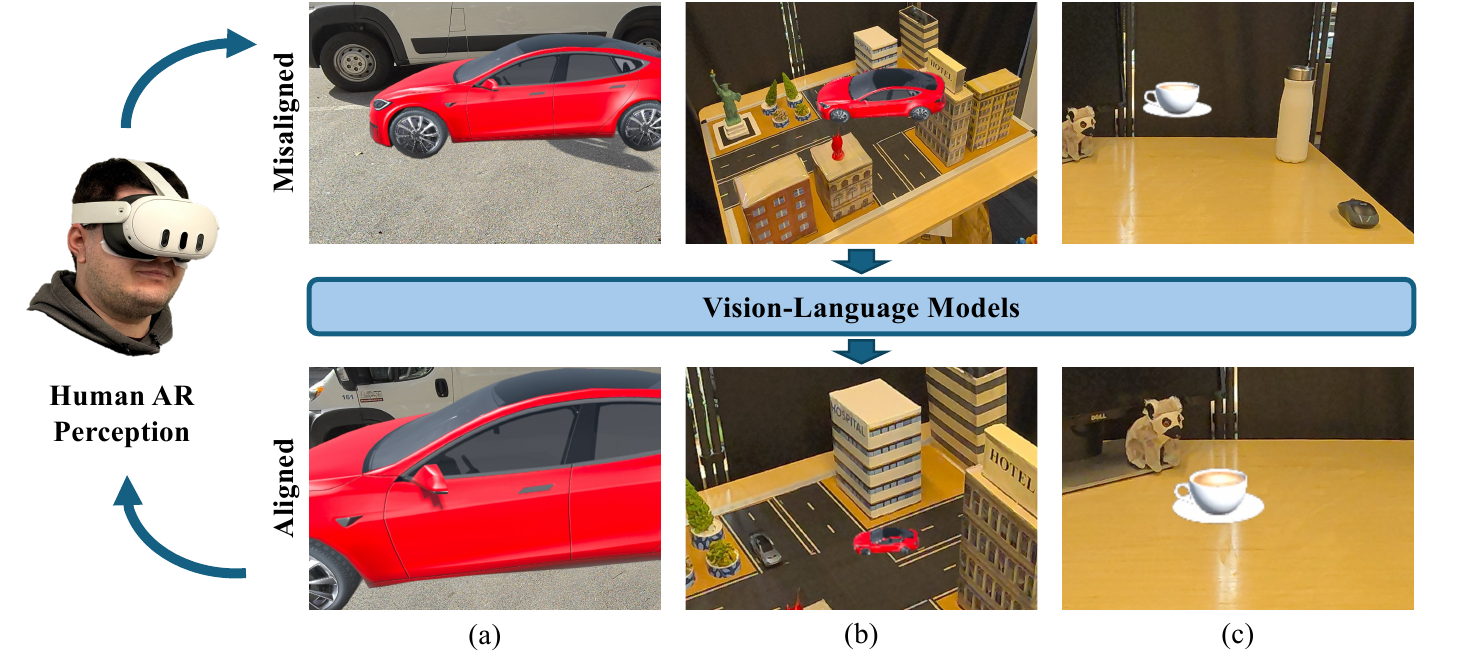}
  \caption{Augmented reality content adjustments guided by vision-language models: (a) a virtual car is scaled to fill a parking spot; (b) a digital car is downsized to fit a toy city's scale; and (c) a coffee cup asset is lowered to rest on the table's surface.}
  \Description{Examples of virtual content being adjusted automatically using vision language models to meet human expectations in augmented reality: (a) a virtual car is scaled to fill a parking spot; (b) a digital car is downsized to fit a toy city's scale; and (c) a coffee cup asset is lowered to rest on the table's surface.}
  \label{fig:teaser}
\end{teaserfigure}

\maketitle

\section{Introduction}

\begin{figure}[hbtp]
\centering
\includegraphics[width=0.47\textwidth]{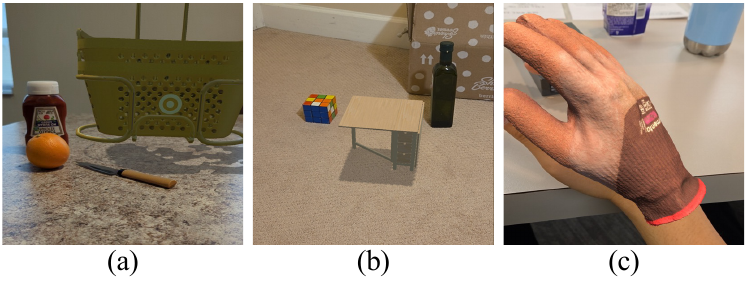} 
\caption[AR deployment challenges from the DiverseAR+ dataset demonstrating conflicts between asset shadow, size, and placement renderings in real-world settings.]{AR deployment challenges from the DiverseAR+ dataset~\cite{duan2025probing}: (a) conflicting real/virtual shadow directions; (b) an unnaturally small AR table; and (c) a spatially misaligned digital glove on the user's hand.}
\Description{AR deployment challenges from the DiverseAR+ dataset demonstrating conflicts between asset shadow, size, and placement renderings in real-world settings.}
\label{figure:AR_deploy_issues}
\end{figure}

Augmented reality (AR) advancements spark innovation, enabling a wide range of applications in education~\cite{chang2022ten}, healthcare~\cite{kanschik2023virtual}, and workplace safety~\cite{chen2024augmented}. Despite rapid developments in AR rendering, inconsistent 3D reconstruction and scene geometry discrepancies on head-mounted displays (HMDs) often impede the presentation of virtual content~\cite{billinghurst2021grand,stern2025impact}. Such phenomena yield perceptual challenges such as implausible shadows~\cite{adams2021shedding}, semantically incorrect content scaling~\cite{qian2022scalar}, and misaligned object placement~\cite{stern2025impact} that break user immersion, as shown in Figure~\ref{figure:AR_deploy_issues}.

Addressing visual perception anomalies remains an open challenge within the AR community. While user studies offer valuable feedback, they are labor-intensive and time-consuming, creating bottlenecks for iterative AR design and deployment~\cite{kosara2003thoughts,ashtari2020creating}. These limitations highlight a need for automatic AR content \emph{assessment} and \emph{adjustment} to reduce dependence on full 3D reconstruction and manual tuning. Existing automatic quality metrics~\cite{hore2010image,wang2023exploring,sekhri2024enhancing} require further evaluation to measure their performance on human-centric AR principles.

Vision-language models (VLMs) offer potential solutions for evaluating and interpreting AR from a human perceptual view. Trained on large-scale multimodal data, VLMs excel at visual understanding and contextual reasoning, making them ideal candidates for scalable, human-aligned AR quality assessment. Prior work exhibits promising results, showcasing that VLM-driven applications have explored vision tasks including foreground-background coherence~\cite{xu2025harmonyiqa}, medical CT scan quality rating~\cite{chen2023iqagpt}, and task-detrimental content detection~\cite{xiu2025viddar}. However, perceptual factors known to influence user immersion, such as spatial alignment~\cite{bang2020effects}, size appropriateness~\cite{abdallah2024factors}, and shadow plausibility~\cite{adams2021shedding}, remain underexplored.

In this work, we investigate the use of VLMs to evaluate and improve visual quality features central to maintaining AR user immersion. Our study is guided by the following research questions:

\begin{itemize}
    \item \textbf{RQ1:} How aligned are commercial VLMs' visual quality assessments with human subjective quality ratings?
    \item \textbf{RQ2:} Can an automated, closed-loop pipeline powered by VLMs generate actionable refinement parameters to mitigate AR rendering artifacts and improve immersion?
    \item \textbf{RQ3:} How do different prompting strategies impact a VLM's ability to reason about AR visual features?
\end{itemize}

To address these inquiries, we first introduce \emph{RateAR}, a curated benchmark comprising $321$ AR images and $112$ AR videos captured from real AR renderings in physical environments covering a diverse range of content artifacts and scene scenarios. Subsequently, we assess the performance of eleven commercial VLMs maintained by four artificial intelligence organizations to rate three AR visual factors: placement plausibility, size appropriateness, and shadow realism (RQ1). Our experimental results support strong assessment alignment between subjective human scores and \emph{GPT}-family~\cite{gpto3,gpt5,gpt54} VLM predictions, with Spearman’s rank-order correlation coefficients (SRCC) up to $0.8695$, Pearson linear correlation coefficients (PLCC) up to $0.8440$, and Kendall’s rank-order correlation coefficients (KRCC) up to $0.7542$. We further conduct an ablation study on prompting strategies, demonstrating that our contextual prompt design provides effective semantic guidance for VLM-based AR quality evaluation (RQ3). Motivated by our findings, we propose and build a self-improving AR content adjustment framework and conduct a $21$-participant study to validate its utility (RQ2). We observe that users appreciate the automatic system, with over $90\%$ of participants reporting improved AR content placement and size coherence. Our contributions are summarized as follows:

\begin{itemize}
    \item We curate and release the \emph{RateAR} dataset,\footnote{\url{https://github.com/Duke-I3T-Lab/RateAR}\label{footnote:rateardataset}} a benchmark comprising $321$ AR images and $112$ AR videos. It provides a diverse set of samples with varying AR content visual quality across three key dimensions: placement plausibility, size appropriateness, and shadow realism.
    \item We compare the performance of eleven commercial VLMs against fine-tuned image and video quality assessment baselines on \emph{RateAR}. Our results demonstrate that the GPT models' ratings consistently exhibit moderate to high correlation with human annotator scores while also numerically outperforming baseline approaches.
    \item We design and conduct a user study with $21$ participants to assess a VLM-based, self-improving AR content adjustment framework. We demonstrate that our pipeline improves AR content placement and size coherence, while $90\%$ of participants agree that our pipeline improves virtual content integration.
\end{itemize}

\section{Related Work}

\subsection{AR Content Perceptual Quality Assessment}

The perceptual quality of virtual content plays an important role in how naturally AR content integrates with the physical environment. Prior work has examined a range of visual factors that influence such integration through controlled human-subject studies. Benjamin \textit{et al.}~\cite{benjamin2024evaluating} examined perceptual effects involving the size, distance, and motion of virtual entities in outdoor AR. Chiu \textit{et al.}~\cite{chiu2025comparing} further studied absolute size perception in AR through verbal and physical judgments. Spatial inconsistencies have also been shown to affect AR perception and interaction: Stern \textit{et al.}~\cite{stern2025impact} investigated the effects of spatial misalignment on collaborative AR experiences, while Au \textit{et al.}~\cite{au2026depth} demonstrated that conflicts between real and virtual depth cues can systematically distort spatial judgments during visually guided interaction. Beyond object geometry, Adams \textit{et al.}~\cite{adams2021shedding} investigated how cast shadows affect perceived ground contact of virtual objects in AR and VR, while Hu \textit{et al.}~\cite{hu2024perception} examined how occlusion handling influences the perceptual integration of real and virtual content. However, these factors have typically been evaluated through controlled user studies or perceptual judgment tasks, requiring participant recruitment and repeated subjective assessment. This reliance on manual evaluation limits scalability for large-scale or iterative AR content assessment.

\subsection{Automatic AR Content Quality Evaluation}

To reduce reliance on repeated human evaluation, prior work has explored automatic methods for AR content quality assessment. Early work adapted full-reference image quality assessment (IQA) by comparing AR renderings to reference images using local metrics such as structural similarity index measure~\cite{hore2010image} and visual information fidelity~\cite{sheikh2006image}. More recent approaches leverage deep learning to better capture perceptual effects, including confusion-theory-based models with attention mechanisms and saliency prediction~\cite{duan2022confusing}, consistency-training frameworks~\cite{eybposh2024coniqa}, and benchmarking studies on AR HMDs~\cite{wang2024subjective}. Despite these advances, existing evaluators often fall short of robust performance across diverse platforms and scenes and typically produce a single holistic score or target a narrow feature, limiting their ability to assess multiple perceptual dimensions. These limitations motivate us to explore the semantic and visual reasoning capabilities of VLMs for factor-specific AR content quality assessment.

\subsection{VLMs for Perceptual Reasoning in AR}

VLMs, such as Gemini 2.5 Pro~\cite{gemini25} and GPT-5~\cite{gpt5}, exhibit remarkable performance across numerous perceptual tasks, including language grounding~\cite{gao2024physically}, scene understanding~\cite{zheng2025video}, and visual question answering~\cite{kuang2025natural}. Recent works further investigate the perceptual reasoning abilities of VLMs. For instance, TopoPerception evaluates global visual perception using topology-based tasks~\cite{zhou2025topoperception}. Similarly, the VQualA challenge introduces large-scale benchmarks requiring multimodal models to reason about subtle visual quality differences across images~\cite{zhu2025vquala}. These perceptual reasoning capabilities have also motivated the increasing adoption of VLMs in AR applications. Existing efforts to integrate VLMs into AR applications have largely focused on using them for virtual content generation and placement. In text-to-3D frameworks, VLMs synthesize digital assets from natural language prompts~\cite{xiu2025say}. Srinidhi \textit{et al.} demonstrate using VLMs to identify physically plausible and semantically appropriate locations for virtual object placement~\cite{srinidhi2024xair}. However, these applications primarily focus on content generation and placement rather than perceptual quality assessment. In a preliminary study, Duan \textit{et al.} provide a limited exploration of VLMs on AR-generated scenes~\cite{duan2025advancing}, largely restricted to coarse scene-level judgments. Their findings indicate that VLMs may succeed on visually conspicuous virtual objects but often fail when virtual content is contextually integrated. However, this work focused on global scene-level assessments without delving into detailed, factor-specific evaluations that reflect nuanced aspects of user-perceived AR quality. Motivated by the perceptual reasoning capabilities of VLMs and their emerging applications in AR, our work investigates their use for factor-specific AR content quality assessment.

\section{RateAR Dataset}\label{sec:RateAR_dataset}

We introduce \textbf{\emph{RateAR}}, a new benchmark supporting feature-level AR content evaluation across diverse environments. The dataset is publicly available on GitHub.\textsuperscript{\ref{footnote:rateardataset}}

\begin{figure}[t]
\centering
\includegraphics[width=0.43\textwidth]{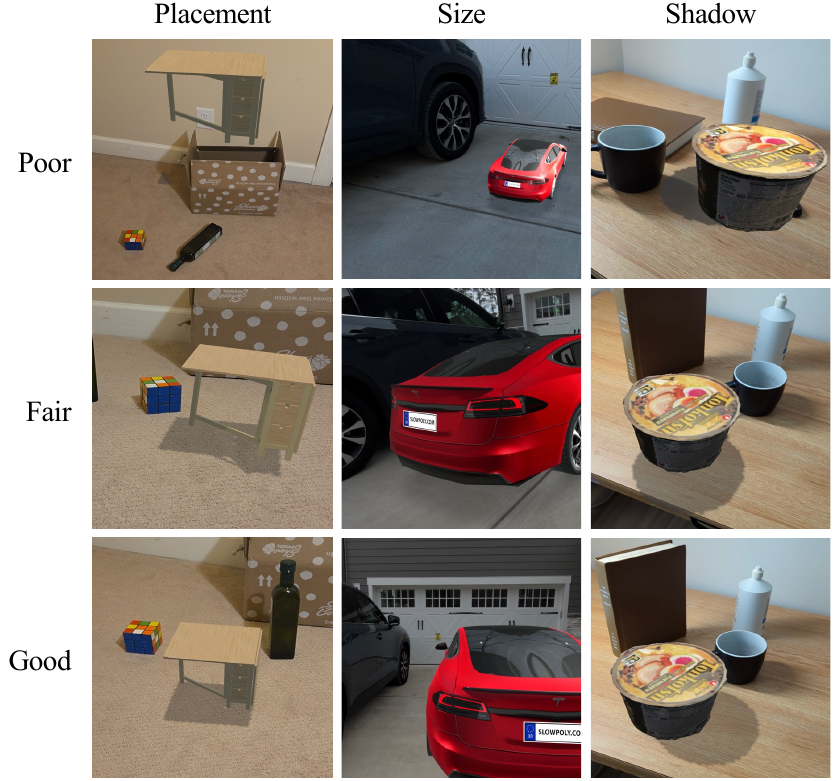} 
\caption[Representative examples from the RateAR dataset illustrating various quality levels across three key visual features: placement plausibility, size appropriateness, and shadow realism.]{\emph{RateAR} dataset examples illustrating varying quality across three features: (a) \emph{Placement plausibility:} virtual desk floating a few feet above (poor), inches above (fair), or level with the ground (good); (b) \emph{Size appropriateness:} digital car appearing far too small (poor), slightly small (fair), or realistically sized (good); and (c) \emph{Shadow realism:} AR noodle bowl shadow opposing the real lighting (poor), slightly misaligned (fair), or fully aligned (good).}
\Description{Representative examples from the RateAR dataset illustrating various quality levels across three key visual features: placement plausibility, size appropriateness, and shadow realism.}
\label{figure:rateARdataset}
\end{figure}

\paragraph{Dataset Composition} 
\emph{RateAR} comprises $321$ AR images and $112$ AR videos, curated to diversify the visual quality levels among influential user immersion properties. Figure~\ref{figure:rateARdataset} presents representative examples from \emph{RateAR}, illustrating a range of quality levels across three key visual factors: placement plausibility, size appropriateness, and shadow realism.

\paragraph{AR Platforms} 
AR samples are sourced from a range of devices and scenarios. $321$ AR images are collected from \textit{DiverseAR+}~\cite{duan2025probing}, selected for their varying rendering qualities in virtual content placement, shadow, and size. All \emph{RateAR} image instances were captured using Android smartphones ($250$), Apple Vision Pro ($40$), and Microsoft HoloLens 2 ($31$). A diverse collection of environments is featured, including bedrooms, kitchens, living rooms, medical offices, reading rooms, research labs, and study rooms. The $112$ AR videos were captured by us to represent a similar distribution of quality feature variations, showcasing settings such as basements, bedrooms, living rooms, and research labs. All \emph{RateAR} videos are recorded using Meta Quest 3.

\paragraph{Context-Dependent Scenarios} 
Alongside common settings such as bedrooms and kitchens, where assessing visual factors primarily relies on general, publicly shared knowledge, \emph{RateAR} also includes scenes that require context-dependent reasoning for evaluation. For example, we include scenarios that capture dynamic user interactions, such as a hand bouncing a virtual basketball. As a result, some examples with observed floating AR content may be contextually appropriate and therefore receive a high placement plausibility score.

\paragraph{Human Subjective Quality Score Labeling}
Four graduate students with experience designing, developing, and evaluating AR systems rated placement plausibility, size appropriateness, and shadow realism using a Python notebook GUI. Before labeling samples, the annotators discussed the target quality features and expectations for handling nuanced edge cases, such as what score to assign when a virtual object without an accompanying shadow was visually appropriate (e.g., anatomical models). As needed, subsequent consensus meetings resolved annotator disagreements. Modeled after the International Telecommunication Union's (ITU) Absolute Category Rating guidelines \cite{international1996methods}, a $5$-point discrete scale was employed. An assignment of $1$ signifies \textit{poorly} rendered feature quality, while a score of $5$ indicates \textit{good} or exceptional visual appearance. After the initial annotation period, we recruited a fifth student with a background in image analysis applications to expand the label set to five ratings per \emph{RateAR} sample. The final annotator received reference materials summarizing the earlier group's discussions to align evaluation standards (made publicly available\textsuperscript{\ref{footnote:rateardataset}}). All annotators assigned quality labels for each AR content factor, per sample, resulting in $6,495$ annotations. The mean opinion scores (MOSs), as computed in~\cite{series2012methodology,duan2022confusing}, are used as final ratings in our evaluations (see \Cref{sec:NewEvaluation}).

To measure inter-annotator reliability, we computed the intraclass correlation coefficients (ICC) for each labeled quality feature based on a two-way random-effects model, average measures, and absolute agreement~\cite{shrout_intraclass_1979,Koo_2016_jcm_15_2} using the \texttt{Pingouin} open-source Python package~\cite{vallat2018pingouin}. The model for placement plausibility yielded $ICC(2,5)=.95$, 95\% $CI[.94,.96]$, $F(432,1728) =22.20$, $p<.001$, supporting excellent reliability. Similarly, we observe excellent reliability for the size appropriateness labels, with $ICC(2,5)=.94$, 95\% $CI[.93,.95]$, $F(432,1728)=16.55$, $p<.001$. Lastly, the shadow realism annotation analysis returns $ICC(2,5)=.90$, 95\% $CI[.87,.92]$, $F(432,1728)=11.15$, $p<.001$, indicating good to excellent reliability.

\section{Methodology}
We investigate the ability of VLMs to evaluate factor-specific quality in virtual content, benchmarking their perception and interpretation against human judgment (RQ1), including during user interactions with AR applications (RQ2). To this end, we (1) design tailored VLM prompts and (2) propose a closed-loop AR content adjustment framework based on VLM evaluations.

\subsection{Prompt Design}
\label{sec:Prompt Template Design}

We design visual feature quality assessment prompts to capture human-perceived contextual semantics, enabling human-aligned evaluations of AR visual quality. To avoid cross-factor interference, we assess each factor independently. For each sample (AR image or video), we provide the VLM with the corresponding prompt and visual input via an API call. A post-processing module parses the VLM’s response to produce the final quality score.

\paragraph{Placement Assessment Prompt} 
Most real-world objects, and therefore AR content, are subject to gravity and should rest on level surfaces. However, certain objects (e.g., helium-filled balloons) or objects captured during dynamic motion (e.g., a bouncing ball) may legitimately appear suspended mid-air in still images. Hence, we design the placement plausibility prompt to consider commonsense scenarios where mid-air objects are acceptable and distinguish them from abnormal floating behavior that results in poor AR content integration with the scene. The prompt is as follows:

\begin{quote}
\noindent\emph{``Please assess whether the virtual object in this AR scene is correctly placed or appears to be floating/misaligned. If it does float or look misaligned, indicate whether that behavior is appropriate for this object type. Rate placement appropriateness from 1 to 5, where 1 means an object that should be grounded/ aligned is clearly floating or misaligned, and 5 means a non-floating object is firmly grounded and properly aligned, or a float-capable object is reasonably floating. Use cues such as whether any rendered shadow (if present) is attached under the object, whether the object is intended for conceptual visualization, and whether the object interacts with forces or other objects (e.g., a ball bouncing under a hand). Please end with a final sentence in this exact format: Score: [your rating].''}
\end{quote}

\paragraph{Size Assessment Prompt}
In most situations, users expect virtual objects' size to approximate that of the objects they represent. However, AR application purposes and scenarios can alter these assumptions. Acknowledging these edge cases, we craft the size appropriateness prompt to consider the AR scene's context and reason whether virtual content size deviations are justifiable. Specifically, the size assessment prompt is:

\begin{quote}
\noindent\emph{``Please assess whether the virtual object’s size appears normal for any plausible subtype of its category (e.g., a `table' could be a coffee table, console/sofa table, or desk) within this AR scene. If not, explain whether the atypical size is appropriate given the surrounding context. Rate size appropriateness on a 1–5 scale, where 1 means an object that should be normal-sized is clearly too large or too small, and 5 means the object is within a normal range or, if intentionally non-standard, is reasonably larger or smaller. Use cues such as nearby reference objects, human-scale affordances, and contextual clues (e.g., a surreal museum). Please end with a final sentence in this exact format: Score: [your rating].''}
\end{quote}

\paragraph{Shadow Assessment Prompt}
AR content shadows contribute differently to overall user immersion across AR applications. As a result, we expect VLMs to reason about each scenario and contextualize the importance of realistic shadows. For example, in many AR medical contexts, attributes such as anatomically accurate placement and size heavily outweigh the necessity of shadows. The prompt is as follows:

\begin{quote}
\noindent\emph{``Please assess whether the virtual object’s shadow is appropriate for this AR scene. If not, explain whether the inconsistency is appropriate given the surrounding context. Rate shadow appropriateness on a 1–5 scale, where 1 means an object that should have a proper shadow instead has a clearly missing, misaligned, or inconsistent one, and 5 means the object has a relatively consistent shadow or, if intentionally non-standard, its absence is reasonable. Use cues such as whether the virtual shadow relatively aligns with nearby reference objects in direction, softness, and intensity, and whether the object is intended for conceptual visualization. Please end with a final sentence in this exact format: Score: [your rating].''}
\end{quote}

\subsection{Autonomous Content Adjustment System}
\label{adjust_framework}

\begin{figure*}[htpb]
\centering
\includegraphics[width=0.87\textwidth]{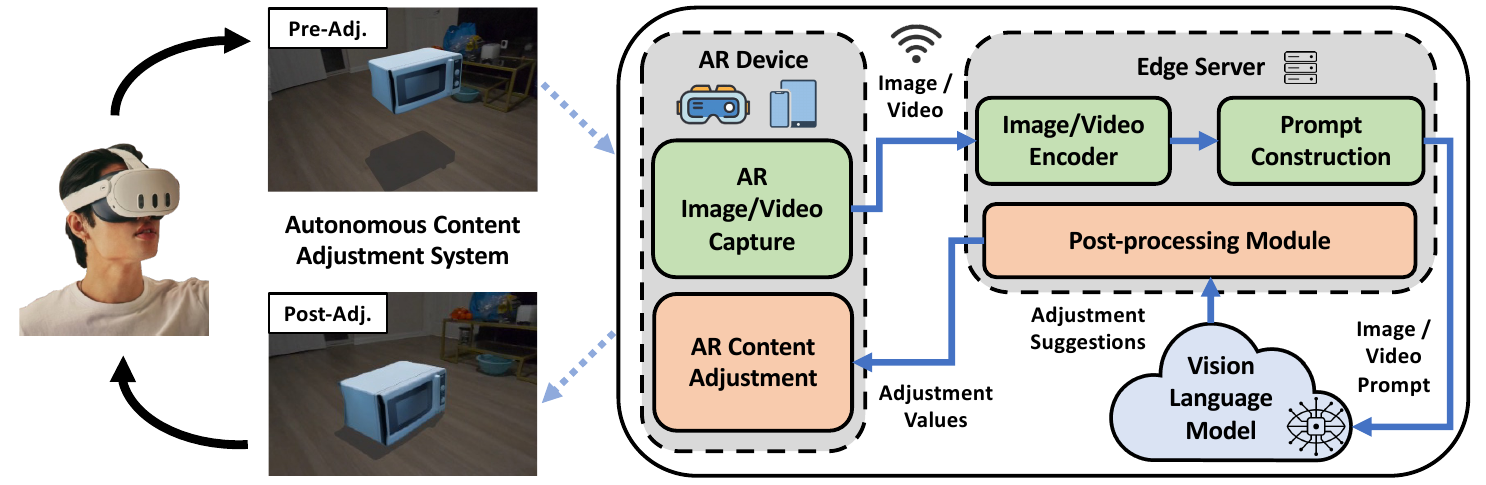}
\caption[Autonomous AR content adjustment framework: The AR device captures an image or video of the scene and sends it to the edge server. The edge server encodes it in base64 format along with an adjustment-aware prompt, then sends the task to a cloud VLM. After inference, the edge server extracts the VLM's suggestions to identify actionable adjustment values within the rendering framework. These changes are sent to the AR device to manipulate the AR content's visual appearance before rerendering.]{Autonomous AR content adjustment framework: The user initially sees a floating virtual microwave (top). Our system evaluates the scene and automatically lowers it to the floor (bottom).}
\Description{Autonomous AR content adjustment framework: The AR device captures an image or video of the scene and sends it to the edge server. The edge server encodes it in base64 format along with an adjustment-aware prompt, then sends the task to a cloud VLM. After inference, the edge server extracts the VLM's suggestions to identify actionable adjustment values within the rendering framework. These changes are sent to the AR device to manipulate the AR content's visual appearance before rerendering.}
\label{figure:adjustframe}
\end{figure*}

One use case for automatic, human-aligned quality assessments in AR applications is correcting virtual content initially rendered poorly within a scene. We propose a VLM-based, closed-loop AR content adjustment framework (see Figure~\ref{figure:adjustframe}) comprising three stages: (1) visual media capture and paired input prompting; (2) quality prediction using commercial VLM APIs; and (3) retrieval of adjustment parameters before applying recommended content rendering suggestions, if any. This automated, visual feedback workflow supports efficient and robust AR quality assessment that aligns well with human perceptual preferences. Additionally, the pipeline can adjust to diverse deployment environments, including edge-cloud infrastructure. 

\paragraph{Data Capture and Prompting} 
Using a compatible AR device, such as a HMD, a single AR image or short video clip is captured and shared with the edge server. If opting for the video-based approach, the initial clip is uniformly subsampled and transformed into an image series. Next, the visual media is Base64-encoded on the edge server for transmission purposes. Finally, the AR input is paired with a context-aware prompt and sent to the VLM for evaluation. The employed prompts (provided in~\Cref{sec:cont_adj_prompts}) expand those introduced in~\Cref{sec:Prompt Template Design}, adding instructions for the VLM to output actionable rendering parameter recommendations. 

\paragraph{VLM Estimation (Zero-Shot)} 
The proposed pipeline utilizes an off-the-shelf commercial VLM through API calls. Each query contains the relevant AR media and prompt. To avoid cross-feature interference, separate VLM API calls are made for each feature quality assessment. We instruct the VLM to return its evaluation output using natural language, specifically asking it to: (1) clearly state its rating; (2) justify its reasoning for the rating; and (3) describe in detail how to improve the AR content's quality, if applicable.

\paragraph{Parameter Extraction and Adjustment}
VLM API calls return outputs to the edge server. Accounting for potentially lengthy replies, a post-processing module extracts the numerical adjustment values. The VLM’s recommendations are then mapped to supported system parameters, resulting in closed-loop refinements without human intervention. 

\section{Experiments}
\label{sec:NewEvaluation}

We evaluate eleven VLMs' abilities to assess AR visual quality and compare correlation results with subjective human MOSs against multiple baseline methods using our \emph{RateAR} benchmark. Subsequently, we describe the user study validating the proposed AR content adjustment framework introduced in Section~\ref{adjust_framework}.

\subsection{Experimental Setting} \label{sec:main_exp_setup}
\paragraph{Evaluated VLMs} 
We evaluate eleven VLMs across four widely used model families: 
Gemini 2.5 Flash, Gemini 2.5 Pro, Gemini 3.5 Flash, and Gemini 3.1 Pro by Google DeepMind~\cite{gemini25,gemini31pro,gemini35flash},
GPT-o3, GPT-5, GPT-5.4-Mini, and GPT-5.4 by OpenAI~\cite{gpto3,gpt5,gpt54},
Grok-4.3 and Grok-4.5 by SpaceXAI~\cite{grok43,grok45},
and Muse Spark 1.1 by Meta Platforms~\cite{museSpark11}. We performed VLM inference using public APIs.

\paragraph{Baselines} 
We compare VLM quality evaluations against the following baselines: (1) traditional and state-of-the-art (SOTA) \textit{image-based} assessments.
Specifically, 
HyperIQA~\cite{su2020blindly},
CNNIQA~\cite{kang2014convolutional},
CLIPIQA+~\cite{wang2023exploring},
LIQE~\cite{zhang2023liqe},
LPIPS-VGG+~\cite{zhang2018unreasonable,chen2024topiq}, 
TOPIQ~\cite{chen2024topiq}, and
QualiCLIP+~\cite{agnolucci2024quality}; and (2) SOTA \textit{video-based} evaluations: 
DOVER~\cite{wu2023exploring}, 
COVER~\cite{he2024cover},
ModularBVQA~\cite{wen2024mbvqa},
ReLaX-VQA~\cite{wang2025relaxvqa},
E-VQA~\cite{sun2025evqa}, and
DIVA-VQA~\cite{wang2025divavqa}. We fine-tune all baselines on \emph{RateAR} from published model weights using an NVIDIA GeForce RTX 3090 GPU, with \texttt{IQA-PyTorch}~\cite{pyiqa} for IQA models and the authors' released training pipelines for video-based models. Additionally, as the baseline frameworks routinely predict a single, summative metric, we train three separate instances of each architecture to evaluate performance on the three feature quality tasks (placement plausibility, size appropriateness, and shadow realism).

\paragraph{Training and Test Set Splitting.}  
To fine-tune the image- and video-based baselines, we create a training set of $100$ AR images and another training set of $50$ AR videos from \emph{RateAR}. Both splits are curated using random samples that are confirmed to include a wide range of context-dependent scenarios. The remaining $221$ AR images and $62$ AR videos form the respective test sets for evaluation. The contextual prompts were developed independently of the test sets, although some generic semantic concepts illustrated in the prompts may occur in both the training and test sets. We evaluate VLMs using zero-shot inference on the test sets, relying on prompt design to align their prior reasoning and semantic knowledge to the \emph{RateAR} quality assessment tasks. 

\paragraph{Evaluation Metrics}
We measure the performance of the VLMs and fine-tuned baseline methods using correlation coefficients. Specifically, we employ SRCC, PLCC, and KRCC. These metrics assess the agreement and consistency between the subjective human ratings and model predictions for feature quality evaluation. For interpretation, correlations below $0.20$ are considered very weak; between $0.20$ and $0.39$, weak; $0.40$ and $0.59$, moderate; $0.60$ and $0.79$, strong; and above $0.80$, very strong~\cite{papageorgiou2022correlation}.

\paragraph{Subjective Score Preprocessing} We preprocessed the subjective scores following the recommendations described in~\cite{series2012methodology}. First, we transformed the raw labels into Z-scores, clipping them to [$-3,+3$] if necessary. Next, we linearly scaled them into the range [$0, 100$]. Finally, we averaged the scaled ratings to compute the MOS for each \emph{RateAR} sample. The associated formulas are as follows:

\vspace{3pt}
\noindent\begin{minipage}{.49\linewidth}
  \begin{equation}
    z_{i,j} = \frac{s_{i,j} - \mu_i}{\sigma_i}
  \end{equation}
\end{minipage}%
,\begin{minipage}{.5\linewidth}
  \begin{equation}
    z^\prime_{i,j} = \frac{100(z_{i,j} + 3)}{6}
  \end{equation}
\end{minipage}

\noindent\begin{minipage}{\linewidth}
\begin{equation}
    MOS_j = \dfrac{1}{N} \sum^N_{i=1} z^\prime_{i,j}
\end{equation}
\end{minipage}%
\vspace{3pt}

\noindent where $s_{i,j}$ is the raw score given by the $i-$th annotator to the $j-$th sample, $\mu_i$ is the mean score given by annotator $i$ (for the target quality factor), $\sigma_i$ is the standard deviation, and $N$ is the total number of annotators. In total, three MOS values are computed for each image and video in \emph{RateAR}.

\subsection{Performance Analysis}\label{sec:perform_analysis}
To evaluate VLM score prediction alignment with human subjective ratings, we run experiments on \emph{RateAR}'s test sets. \Cref{tab:main_results} and \Cref{tab:video_results} report correlation metrics between automated ratings and human MOSs for the three quality features. Corresponding coefficient $95\%$ confidence intervals and statistical significance p-values are made available in the Supplemental Materials; see 
\Cref{tab:main_results_CI,tab:main_results_pvals,tab:video_results_CI,tab:video_results_pvals}.

\begin{table*}[hbtp] 
\centering
\caption{Correlation metrics between human scores and model predictions on RateAR images. Bold indicates column-wise best.}
\label{tab:main_results}
\begin{tabular}{@{}l|rrr|rrr|rrr}
\toprule
                             \multirow{2}{*}{\textbf{Methods}} & \multicolumn{3}{c|}{\begin{tabular}[c]{@{}c@{}}\textbf{Placement}\end{tabular}}                                    & \multicolumn{3}{c|}{\begin{tabular}[c]{@{}c@{}}\textbf{Size}\end{tabular}}                                      & \multicolumn{3}{c}{\begin{tabular}[c]{@{}c@{}}\textbf{Shadow}\end{tabular}}                                             \\ \cline{2-10} 
    & \multicolumn{1}{c}{SRCC}               & \multicolumn{1}{c}{PLCC}               & \multicolumn{1}{c|}{KRCC}              & \multicolumn{1}{c}{SRCC}               & \multicolumn{1}{c}{PLCC}               & \multicolumn{1}{c|}{KRCC}              & \multicolumn{1}{c}{SRCC}               & \multicolumn{1}{c}{PLCC}               & \multicolumn{1}{c}{KRCC}               \\ \hline
HyperIQA                   & 0.5093                                 & 0.5084                                 & 0.3727                                 & 0.6767                                 & 0.6418                                 & 0.5150                                 & 0.5025                                 & 0.5470                                 & 0.3585                                 \\
CNNIQA                        & 0.3595                                 & 0.3666                                 & 0.2573                                 & 0.4512          & 0.4737                                 & 0.3216                                 & 0.3154                                 & 0.3734          & 0.2219                                 \\
CLIPIQA+                     & 0.3792                                 & 0.3614                                 & 0.2749                                 & 0.6258                                 & 0.5721                                 & 0.4388                                 & 0.4038                                 & 0.4768                                 & 0.2814                                 \\
LIQE                   & 0.4564                                 & 0.4518                                 & 0.3293                                 & 0.7079                                 & 0.6838          & 0.5283          & 0.4950                                 & 0.5413                                 & 0.3525                                 \\
LPIPS-VGG+                     & 0.6970                                 & 0.6504                                 & 0.4955                                 & 0.5672                                 & 0.4934                                 & 0.3913                                 & 0.3441          & 0.3359 & 0.2416                                 \\
TOPIQ                       & 0.5526                                 & 0.5165                                 & 0.3972                                 & 0.6690                                 & 0.5793                                 & 0.4833                                 & 0.5348                                 & 0.5627                                 & 0.3944                                 \\ 
QualiCLIP+                       & 0.4221                                 & 0.4557                                 & 0.2982                                 & 0.6936                                 & 0.6670                                 & 0.5191                                 & 0.4887                                 & 0.5609                                 & 0.3437                                 \\
\hline
Gemini-2.5-Flash                  & 0.5968                                 & 0.5763                                 & 0.4776                                 & 0.2480                                 & 0.2971                                 & 0.1959                                 & 0.4472                                 & 0.4317                                 & 0.3363                                 \\
Gemini-2.5-Pro & 0.5916                                 & 0.5702                                 & 0.4798                                 & 0.3311                                 & 0.3759                                 & 0.2699                                 & 0.3850                                 & 0.3084                                 & 0.3018                                 \\
Gemini-3.5-Flash               & 0.7351                                 & 0.7430                                 & 0.6061                                 & 0.6567                                 & 0.7253                                 & 0.5285                                 & 0.4816                                 & 0.4440                                 & 0.3866                                 \\
Gemini-3.1-Pro                       & 0.4750          & 0.4249          & 0.3920          & 0.4730 & 0.5085 & 0.3739 & 0.3078 & 0.2844                                 & 0.2518 \\
GPT-o3                        & 0.7416 & 0.7154 & 0.6078 & \textbf{0.7378}                                 & \textbf{0.7562}                                 & \textbf{0.5812}                                 & 0.5866                                 & 0.5440                                 & 0.4677          \\
GPT-5               & \textbf{0.8336}                                 & \textbf{0.7979}                                 & \textbf{0.6845}                                 & 0.4968                                 & 0.5832                                 & 0.3873                                 & 0.5025                                 & 0.4927                                 & 0.3961                                 \\
GPT-5.4-Mini                & 0.5695                                 & 0.5524                                 & 0.4582                                 & 0.0436                                 & 0.0886                                 & 0.0341                                 & 0.3395                                 & 0.3190                                 & 0.2645                                 \\
GPT-5.4               & 0.7196                                 & 0.6963                                 & 0.5865                                 & 0.2110                                 & 0.3452                                 & 0.1569                                 & \textbf{0.6075}                                 & 0.5827                                 & \textbf{0.4907}                                 \\
Grok-4.3               & 0.5845                                 & 0.5679                                 & 0.4763                                 & 0.1706                                 & 0.2259                                 & 0.1363                                 & 0.5870                                 & \textbf{0.5875}                                 & 0.4555                                 \\
Grok-4.5                & 0.6121                                 & 0.6135                                 & 0.4892                                 & 0.4836                                 & 0.5638                                 & 0.3905                                 & 0.5984                                 & 0.5790                                 & 0.4560                                 \\
Muse-Spark-1.1                & 0.7836                                 & 0.7832                                 & 0.6462                                 & 0.6730                                 & 0.6947                                 & 0.5372                                 & 0.4563                                 & 0.4740                                 & 0.3624                                 \\
\bottomrule
\end{tabular}
\end{table*}

\begin{table*}[hbtp] 
\centering
\caption{Correlation metrics between human scores and model predictions on RateAR videos. Bold indicates column-wise best.}
\label{tab:video_results}
\begin{tabular}{@{}l|rrr|rrr|rrr}
\toprule
                             \multirow{2}{*}{\textbf{Methods}} & \multicolumn{3}{c|}{\begin{tabular}[c]{@{}c@{}}\textbf{Placement}\end{tabular}}                                    & \multicolumn{3}{c|}{\begin{tabular}[c]{@{}c@{}}\textbf{Size}\end{tabular}}                                      & \multicolumn{3}{c}{\begin{tabular}[c]{@{}c@{}}\textbf{Shadow}\end{tabular}}                                             \\ \cline{2-10} 
    & \multicolumn{1}{c}{SRCC}               & \multicolumn{1}{c}{PLCC}               & \multicolumn{1}{c|}{KRCC}              & \multicolumn{1}{c}{SRCC}               & \multicolumn{1}{c}{PLCC}               & \multicolumn{1}{c|}{KRCC}              & \multicolumn{1}{c}{SRCC}               & \multicolumn{1}{c}{PLCC}               & \multicolumn{1}{c}{KRCC}               \\ \hline
DOVER                     & 0.4444                                 & 0.4372                                 & 0.3206                                 & 0.5084                                 & 0.6189                                 & 0.3549                                 & -0.0334                                & 0.0256                                 & -0.0212                                 \\
COVER                     & 0.3800                                 & 0.4231                                 & 0.2700                                 & 0.4614                                 & 0.5931                                 & 0.3110                                 & 0.0450                                 & -0.0385                                 & 0.0368                                 \\ 
ModularBVQA                     & 0.4348                                 & 0.4391                                 & 0.3126                                 & 0.6348                                 & 0.7756                                 & 0.4450                                 & 0.2190                                 & 0.1438                                 & 0.1451                                 \\
ReLaX-VQA                     & 0.3385                                 & 0.3456                                 & 0.2137                                 & 0.4755                                 & 0.5796                                 & 0.3461                                 & -0.0371                                 & -0.1027                                 & -0.0212                                 \\
E-VQA                     & 0.0830                                 & 0.0937                                 & 0.0563                                 & 0.2790                                 & 0.3412                                 & 0.1890                                 & 0.0162                                 & 0.0860                                 & 0.0112                                 \\
DIVA-VQA                     & 0.3288                                 & 0.3670                                 & 0.2413                                 & 0.5007                                 & 0.6301                                 & 0.3791                                 & 0.0992                                 & 0.0754                                 & 0.0659                                 \\
\hline
Gemini-2.5-Flash                  & 0.6796                                 & 0.6719                                 & 0.5787                                 & 0.2566                                 & 0.3107                                 & 0.2067                                 & 0.3533                                 & 0.3936                                 & 0.2809                                 \\
Gemini-2.5-Pro & 0.4572                                 & 0.4454                                 & 0.3687                                 & 0.3400                                 & 0.4208                                 & 0.2869                                 & 0.4854                                 & 0.4058                                 & 0.3894                                 \\
Gemini-3.5-Flash               & 0.7078                                 & 0.7352                                 & 0.6195                                 & 0.3544                                 & 0.4513                                 & 0.2936                                 & 0.4492                                 & 0.3500                                 & 0.3694                                 \\
Gemini-3.1-Pro                       & 0.4029          & 0.4193          & 0.3523          & 0.1865 & 0.2189 & 0.1525 & 0.1883 & 0.0906                                 & 0.1602 \\
GPT-o3                        & 0.7074 & 0.7759 & 0.5849 & 0.6298                                 & 0.7041                                 & 0.5201                                 & 0.6243                                 & 0.5976                                 & 0.5072          \\
GPT-5               & \textbf{0.8458}                                 & 0.7819                                 & \textbf{0.7058}                                 & \textbf{0.8206}                                 & \textbf{0.8252}                                 & \textbf{0.6865}                                 & \textbf{0.8695}                                 & 0.8440                                 & \textbf{0.7542}                                 \\
GPT-5.4-Mini                & 0.4673                                 & 0.4378                                 & 0.3853                                 & 0.4602                                 & 0.4522                                 & 0.3940                                 & 0.5902                                 & 0.6736                                 & 0.5004                                 \\
GPT-5.4               & 0.5071                                 & 0.5679                                 & 0.4121                                 & 0.4449                                 & 0.4435                                 & 0.3704                                 & 0.8156                                 & 0.8396                                 & 0.7072                                 \\
Grok-4.3               & 0.6443                                 & 0.6338                                 & 0.5536                                 & 0.5752                                 & 0.6204                                 & 0.4871                                 & 0.8248                                 & \textbf{0.9028}                                 & 0.6751                                 \\
Grok-4.5                & 0.7750                                 & 0.7013                                 & 0.6497                                 & 0.3991                                 & 0.4552                                 & 0.3305                                 & 0.7493                                 & 0.7955                                 & 0.6336                                 \\
Muse-Spark-1.1                & 0.8015                                 & \textbf{0.7952}                                 & 0.6916                                 & 0.5026                                 & 0.5662                                 & 0.4147                                 & 0.4575                                 & 0.4846                                 & 0.3813                                 \\ \bottomrule
\end{tabular}
\end{table*}

\subsubsection{Image-based Assessment} 
\Cref{tab:main_results} shows two trends:

(1) \textit{VLMs vs. Baselines.} Across all three factors, GPT-flagship models, excluding GPT-5.4-Mini, achieve the strongest numerical correlations with human ratings, outperforming the best baseline by $0.03\sim0.14$, $0.02\sim0.15$, and $0.05\sim0.19$ in SRCC, PLCC, and KRCC, respectively. This suggests mainline GPT models may better capture context-specific cues. For example, they recognize that balloons can plausibly float and thus assign higher placement plausibility scores to floating balloons. Similarly, they infer that a ``car'' appearing in an office setting is likely a small toy, resulting in higher size appropriateness scores for a small car. Representative textual justifications from the VLMs are available in our GitHub repository.\textsuperscript{\ref{footnote:rateardataset}} In contrast, supervised baselines, fine-tuned on a limited (albeit diverse) set, tend to overfit factor-specific features and struggle to generalize to novel backgrounds and factor states. Other VLMs generally underperform the GPT models, with only a marginal $0.005$ improvement in PLCC for the shadow realism factor, possibly due to less exposure to AR imagery in their pretraining corpora.

(2) \textit{Differences across Features.} For the GPT models, placement plausibility exhibits the highest correlations (e.g., up to $0.8336$ in SRCC), while shadow realism is the most challenging factor (e.g., strongest SRCC is $0.6075$); size appropriateness falls in between. Placement plausibility often hinges on relatively unambiguous geometric cues (e.g., object–support contact) and widely shared commonsense exceptions (e.g., balloons can float, and mid-air sport actions can momentarily suspend objects). By contrast, shadow realism depends on subtler photometric attributes (shadow strength, direction, softness) whose importance varies by context and rater, yielding lower correlations. Although guided by broad human priors, the context dependency of size appropriateness produces slightly lower correlations than placement plausibility.

Overall, these results indicate strong alignment between VLMs' assessments and human judgments, supporting the feasibility of using VLMs as scalable, automated, human-aligned tools to assist AR content assessment and adjustment.

\subsubsection{Video-based Assessment} 
\Cref{tab:video_results} highlights three findings:

(1) \textit{VLMs vs. Baselines.} Across the three factors, GPT-5 numerically achieves the strongest SRCC and KRCC correlations, surpassing the best baseline by $0.19\sim0.65$ and $0.24\sim0.61$, respectively. For PLCC, Muse-Spark-1.1 and Grok-4.3 obtain the highest correlations for placement plausibility and shadow realism, respectively, while GPT-5 performs best for size appropriateness. This further validates the superior context-aware, human-aligned AR scene assessment capabilities of VLMs, particularly the GPT models.

(2) \textit{Differences across Features.} In contrast to image-based trends, shadow realism correlated most strongly across all three features, exceeding placement plausibility by $0.02$ SRCC, $0.11$ PLCC, and $0.05$ KRCC. This reversal likely stems from limited shadow variability in our video set (primarily two states: correct shadow and no shadow), simplifying the shadow realism rating task.

(3) \textit{Videos vs. Images.} The strongest video VLMs numerically outperform the leading image VLMs in human correlation across all features, improving SRCC, PLCC, and KRCC by up to $0.26$, $0.32$, and $0.26$, respectively. This is likely because videos provide temporal cues and multiple viewpoints unavailable in single-frame captures.

Overall, these results indicate that videos provide richer evidence for AR quality assessment, and that VLMs can accelerate AR design loops via scalable, human-aligned evaluations.

\subsection{User Study Design} \label{sec:user_study_design}
Following validation of the VLMs’ visual quality assessment capabilities, we further evaluate the proposed framework through a Duke University IRB-approved user study encompassing both standard and context-dependent scenarios.  

\subsubsection{System Implementation} 
The custom-built application is developed with Unity 2022.3.28f1 and deployed on a Meta Quest 3 headset with accompanying Meta Quest Touch Plus controllers. Default device display settings were employed for all experiments. A MacBook Air (M1) laptop serves as the edge server. Communication between the headset and edge server occurs over one-hop 5~GHz WiFi (802.11n). The cloud server is accessed through OpenAI's API, providing service to the GPT-o3 VLM.

\subsubsection{Participants}
We recruited $21$ participants via on-campus advertisements ($6$ female, $13$ male, $2$ preferred not to say; mean age: $25.6$, range: $20-31$). Regarding familiarity with AR devices, 14.3\% had no prior AR experience, 33.3\% had used AR once or twice, 23.8\% used AR infrequently, and 28.6\% used AR frequently. 57.1\% had played AR games before, and 42.9\% had prior AR development experience.

\subsubsection{User Study Scenarios} 
\begin{figure}[tb]
\centering
\includegraphics[width=0.47\textwidth ]{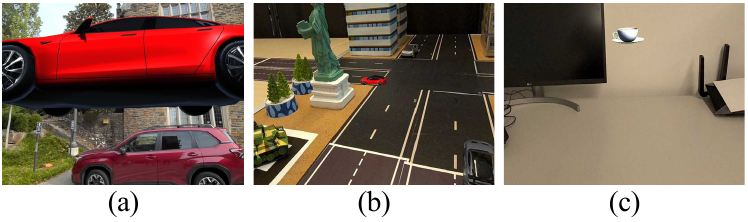}
\caption[User study scenarios utilized throughout framework experimentation: parking lot, miniature toy city, and office desk.]{User study scenarios: (a) Parking lot, for scaling a virtual car relative to parking spaces; (b) Toy city, for rendering the car among miniature buildings; and (c) Office table, for desktop item placement and sizing.}
\Description{User study scenarios utilized throughout framework experimentation: parking lot, miniature toy city, and office desk.}
\label{fig:userstudy_scenario}
\end{figure}

To evaluate the performance and generalization ability of the self-improving AR content adjustment framework, the user study includes three representative scenarios and four types of AR objects. The scenarios include: (1) a \textit{parking lot}, where a randomly scaled or misplaced virtual car is assessed and adjusted by the VLM, aiming to achieve realistic placement and scaling relative to real-world parking spaces; (2) a \textit{toy city}, where a randomly scaled or misplaced virtual car is evaluated and adjusted by the VLM to ensure that the vehicle is placed and sized appropriately relative to the surrounding miniature structures; and (3) an \textit{office table}, which examines the placement and scaling of common desktop items (e.g., a coffee cup and a cardboard box) as well as a virtual inflated balloon used to evaluate physically plausible ``floating” behavior above the table surface. These scenarios represent diverse AR deployment conditions, including large-scale outdoor environments, miniature contexts requiring fine-grained scale reasoning, and tabletop scenes involving both contact-based placement and physically plausible floating objects. Figure~\ref{fig:userstudy_scenario} provides a snapshot of sample user study scenarios. Indoor trials were conducted under controlled laboratory lighting, while outdoor trials were conducted in a parking lot under sunny or partially overcast conditions with at least two researchers supervising for safety.

\subsubsection{Study Procedure} All participants provided written informed consent and received a study overview and quality guide (available on GitHub\textsuperscript{\ref{footnote:rateardataset}}). Individuals acknowledged they could withdraw at any time or request that their data be excluded from subsequent analyses, even after trial completion. Participants were instructed via audio prompts to evaluate either the placement plausibility or the size appropriateness of virtual objects within given AR scenarios, providing ratings both before and after VLM-driven adjustments while wearing the head-mounted display system. The shadow realism quality feature was not analyzed by participants; it is discussed further in~\Cref{sec:limits_and_future}.

Each participant was randomly assigned two virtual objects: one for a \textit{placement-adjustment} task and one for a \textit{size-adjustment} task, following a counterbalanced assignment scheme designed to distribute object types and scenarios evenly across participants. For each object, participants completed two trials: (1) an image-based trial (single, limited-viewport capture); and (2) a 10-second video-based trial (short clip, supporting multi-angle exploration). During each trial, participants wore a Meta Quest 3 headset and launched the app to visualize the assigned virtual object in the AR scene. Participants first provided an initial (pre-adjustment) quality rating. Using a touch controller, subjects then pressed a button to trigger the system to capture an image or a video of the AR scene and query the VLM for adjustment recommendations. After applying the rendering adjustment, when necessary, participants provided a post-adjustment quality rating. Upon completing all four trials, participants filled out a brief survey. Each session lasted approximately 15 to 20 minutes.

\subsubsection{Survey Questionnaire}
Participants completed both pre- and post-experiment surveys.\textsuperscript{\ref{footnote:rateardataset}} In the pre-experiment survey, we collected demographic information and asked questions about their prior experience with AR technologies and AR content design to characterize participants’ background and expertise. The post-experiment survey consisted of questions designed to gather feedback on the system and the overall user experience. For the interaction experience, participants were asked whether the adjustment process felt sufficiently fast during the task. For the system effectiveness, participants were asked whether they found the adjustment system helpful overall and whether they preferred video-based trials over image-based trials when reviewing the adjustment process. All questions in these categories were answered on a five-point Likert scale. Finally, participants were invited to provide open-ended feedback about their overall experience with the system.

\subsection{User Study Results} \label{sec:user_study_results}

\subsubsection{Adjustment Performance} \label{sec:adj_performance} 
We evaluate the framework using pre- and post-VLM adjustment user ratings. Of the 84 recorded trials, we observed improvements in 45 (53.6\%) and no perceived changes in 38 (45.2\%), with only one trial (1.2\%) resulting in worse quality. Many user trials showed no change due to a ceiling effect, where random initialization already yielded maximum quality for both users and the VLM (initial score of $5$ in $23.8\%$, or $20$ of $84$, of cases). Additionally, we observed that the VLM struggled to provide appropriate adjustment instructions when the visual input contained partially occluded virtual AR content. Overall, the results support that VLMs provide human-aligned feedback to correct AR rendering, enabling automated, closed-loop placement and sizing.

Furthermore, we employed two-sided paired Wilcoxon signed-rank tests to evaluate the pre- and post-adjustment score differences. For the image-based placement pipeline, post-modification ratings ($Mdn=3, IQR=2$) were significantly higher than pre-modification ratings ($Mdn=1, IQR=1$; $Z=-3.23, p<.01, r_{c}=-1$, where $r_{c}$ is the matched pairs rank biserial correlation). Similar improvements were observed for the video-based placement pipeline, which increased from a pre-adjustment $Mdn=1$ ($IQR=1$) to a post-adjustment $Mdn=3$ ($IQR=3; Z=-3.24, p<.01, r_{c}=-1$). Regarding size adjustments, the image-based pipeline yielded a significant increase from pre-adjustment ($Mdn=3, IQR=4$) to post-adjustment ($Mdn=4, IQR=2$; $Z=-2.57, p<.05, r_{c}=-.91$). Finally, the video-based size pipeline showed a comparable upward shift from $Mdn=3$ ($IQR=3$) to $Mdn=4$ ($IQR=1; Z=-2.85,$ $ p<.01, r_{c}=-1$). Across all conditions, the large rank biserial correlation magnitudes suggest that our framework improves perceived AR content placement and sizing (see~\Cref{sec:extend_user_study} for p-values).

\subsubsection{Survey Analysis}
We define the \textit{positivity rate} as the proportion of `strongly agree' and `agree' responses. Participants’ free-text responses are reported with the participant number, $P$. \textbf{19/21~participants (90.5\%)} agreed that the self-improving AR content adjustment system increased placement or size quality. Regarding timeliness, 13 of the 21 subjects (61.9\%) were satisfied with the image-based pipeline, compared to only 7 subjects (33.3\%) for the video-based pipeline. This is expected, as processing subsampled video frames for the video-based approach measurably increases recommendation latency. Participants who appreciated the self-improving system described it as `really great and fun' (P1, P3, P17, P21) and commented that it `makes the AR content look more natural' (P5, P9, P10, P18). In summary, user feedback indicates that the system is helpful and sufficiently responsive for interactive use.

\section{Ablation Study}

We conduct ablation experiments to assess the influence of different prompting strategies on VLM-based AR content feature quality evaluation (RQ3). To ensure reliable evaluation, we design separate prompts for each feature, enabling factor-specific judgments while minimizing cross-factor interference.

\begin{table*}[htbp]
\centering
\caption{Correlation metrics between human scores and \emph{GPT-family} flagship model predictions on RateAR images. Bold indicates column-wise best.}
\label{tab:ablation}
\begin{tabular}{@{}ll|rrr|rrr|rrr}
\toprule
                             \multirow{2}{*}{\begin{tabular}[c]{@{}l@{}}\textbf{Prompt}\\ \textbf{Strategy}\end{tabular}} & \multirow{2}{*}{\textbf{VLM}} & \multicolumn{3}{c|}{\begin{tabular}[c]{@{}c@{}}\textbf{Placement}\end{tabular}}                                    & \multicolumn{3}{c|}{\begin{tabular}[c]{@{}c@{}}\textbf{Size}\end{tabular}}                                      & \multicolumn{3}{c}{\begin{tabular}[c]{@{}c@{}}\textbf{Shadow}\end{tabular}}                                             \\ \cline{3-11} 
 &     & \multicolumn{1}{c}{SRCC}               & \multicolumn{1}{c}{PLCC}               & \multicolumn{1}{c|}{KRCC}              & \multicolumn{1}{c}{SRCC}               & \multicolumn{1}{c}{PLCC}               & \multicolumn{1}{c|}{KRCC}              & \multicolumn{1}{c}{SRCC}               & \multicolumn{1}{c}{PLCC}               & \multicolumn{1}{c}{KRCC}               \\ \hline
 & GPT-o3                   & 0.7119                                 & 0.7049                                 & 0.5776                                 & 0.6298                                 & 0.6509                                 & 0.5080                                 & 0.4407                                 & 0.4118                                 & 0.3555                                 \\
Primitive & GPT-5                        & 0.7394                                 & 0.7246                                 & 0.6061                                 & 0.4940          & 0.5364                                 & 0.3776                                 & 0.4686                                 & 0.4638          & 0.3723                                 \\
& GPT-5.4                     & 0.6487                                 & 0.6354                                 & 0.5300                                 & 0.0292                                 & 0.1008                                 & 0.0219                                 & 0.6153                                 & \textbf{0.5984}                                 & \textbf{0.5008}                                 \\ \hline
 & GPT-o3                  & 0.7416                                 & 0.7154                                 & 0.6078                                 & \textbf{0.7378}                                 & \textbf{0.7562}                                 & \textbf{0.5812}                                 & 0.5866                                 & 0.5440                                 & 0.4677                                 \\
Contextual & GPT-5 & \textbf{0.8336}                                 & 0.7979                                & \textbf{0.6845}                                 & 0.4968                                 & 0.5832                                 & 0.3873                                 & 0.5025                                 & 0.4927                                 & 0.3961                                 \\
& GPT-5.4               & 0.7196                                 & 0.6963                                 & 0.5865                                 & 0.2110                                 & 0.3452                                 & 0.1569                                 & 0.6075                                 & 0.5827                                 & 0.4907                                 \\ \hline
 & GPT-o3                  & 0.7222                                 & 0.7202                                 & 0.5841                                 & 0.6560                                 & 0.7221                                 & 0.5183                                 & 0.5449                                 & 0.4947                                 & 0.4257                                 \\
Few-shot & GPT-5 & 0.8009                                 & \textbf{0.8182}                                 & 0.6512                                 & 0.5738                                 & 0.6284                                 & 0.4507                                 & 0.3629                                 & 0.3877                                 & 0.2774                                 \\
& GPT-5.4               & 0.6853                                 & 0.6775                                 & 0.5501                                 & 0.3383                                 & 0.4347                                 & 0.2512                                 & \textbf{0.6258}                                 & 0.5917                                 & 0.4855                                 \\ \bottomrule
\end{tabular}

\end{table*}

\subsection{Prompt Engineering}

We evaluate three prompting strategies, varying in the level of complexity or information provided to the VLM: (1) primitive, (2) contextual, and (3) few-shot prompting.

\paragraph{Primitive Prompts} We design the primitive prompts, which supply the VLM with concise task instructions and no additional details. These prompts are concise but rely heavily on the model's underlying semantic embeddings as they offer no definitions or examples to guide the model's input processing and reasoning. Complete primitive prompts can be found in~\Cref{sec:ablation_primative}.

\paragraph{Contextual Prompts} We curate the contextual prompts by expanding the primitive prompt instructions. They include supporting examples relevant to AR scenarios and aim to provide the models with targeted visual cues that humans perceive. The employed contextual prompts are those introduced in \Cref{sec:Prompt Template Design}.

\paragraph{Few-shot Prompts} This set of queries uses the contextual prompts coupled with the few-shot learning technique~\cite{brown2020language}. Specifically, three instances of \{prompt-image-output\} pairings are presented to the VLM before processing the input. The first example highlights a scene with poorly rendered AR content (score $= 1$), whereas the second depicts exemplar visual quality (score $= 5$). The final example addresses an ambiguous case and presents an AR object whose quality is neither preferred nor terrible (score $= 3$). These illustrations seek to align the model's processing with human perception. Complete few-shot prompts can be found in~\Cref{sec:ablation_few_shot}.

\subsection{Ablation Setup}

We select the \emph{GPT}-family flagship models (GPT-o3, GPT-5, and GPT-5.4) for the ablation evaluation due to their observed higher correlation with human visual perception, as supported by the results reported in \Cref{sec:perform_analysis}. VLM inference is performed through API calls. Again, placement plausibility, size appropriateness, and shadow realism are the target perceptual quality features.   

The \textit{RateAR} image test set ($n=221$), described in \Cref{sec:main_exp_setup}, serves as the ablation benchmark. Primitive and contextual prompt experiments follow zero-shot inference practices. Few-shot prompting experiments source examples from outside the RateAR test set to guide the invoked few-shot inference. Each perceptual quality feature is assigned a dedicated set of few-shot examples for prompting. All trials evaluating the same feature share the same example set, while different features use different sets. In total, three distinct example sets are used, corresponding to the three quality features. SRCC, PLCC, and KRCC remain as the evaluation metrics measuring alignment between human and VLM perception.

\subsection{Prompting Strategy Analysis}

\Cref{tab:ablation} presents the ablation results for the investigated VLM prompting strategies (see \Cref{tab:ablation_CI} and \Cref{tab:ablation_pvals} for coefficient $95\%$ confidence intervals and p-values, respectively). We draw attention to two observations:

(1) \textit{Prompting strategies benefit from guiding cues.} Comparing correlation coefficients across prompting methods, we observe improved alignment with human perception ratings in most experimental trials when prompted with supplemental information. For example, GPT-o3 achieves SRCC, PLCC, and KRCC gains of $0.01\sim0.15$, $0.01\sim0.13$, and $0.01\sim0.11$ for all three factors, respectively, when using the contextual or few-shot prompts compared to the primitive prompts. Although for shadow realism, the highest PLCC/KRCC is obtained with primitive prompts by GPT-5.4, the difference is marginal ($0.01/0.01$) compared to GPT-5.4 with contextual or few-shot prompts. These results suggest that prompting strategies incorporating contextual guidance can better align AR content quality with human perception, as demonstrated by the contextual and few-shot prompting approaches.

(2) \textit{Providing relevant contextual cues in prompts improves evaluation performance, while excessive contextual information can be unnecessary and yield limited additional benefit.} The results in \Cref{tab:ablation} further suggest that while VLMs benefit from semantically crafted prompts, performance gains may plateau or even degrade when models are presented with excessive examples. For instance, contextual prompts yield the strongest VLM-human correlations in 5 of the 9 feature-metric pairings, whereas few-shot prompting leads in only 2. One possible explanation is that the selected few-shot examples may not be sufficiently diverse to represent the full spectrum of feature quality variations. Alternatively, the provided examples may inadvertently constrain the VLMs’ inherent generalization capability. These findings indicate that our contextual prompts provide effective semantic guidance without the limitations of excessive or narrow examples.

Overall, the results suggest that prompt design plays a critical role in VLM-based AR quality evaluation. In particular, our contextual prompts provide effective semantic guidance without the unnecessary complexity introduced by additional examples. 

\section{Limitations and Future Work} \label{sec:limits_and_future}

\paragraph{Dataset Composition} 
\emph{RateAR} combines images selected from \textit{DiverseAR+}~\cite{duan2025probing} and videos collected using Meta Quest 3. Differences in scene composition, media source, and factor realization limit direct image-video comparisons. \emph{RateAR} also covers only a subset of immersion-related visual features. Future work will construct better-matched image-video counterparts and explore additional features, including occlusion and object clipping.

\paragraph{Prototype Scope} The deployed system is a proof-of-concept prototype rather than a full-scale AR stack. The current study evaluates a single VLM-guided adjustment step per trial and operationalizes only placement and size refinement. Although evaluated in our benchmark, shadow realism was omitted from the deployed system due to its lower assessment reliability and the lack of a robust shadow adjustment pipeline for the prototype. In future work, we plan to investigate which shadow attributes are most salient to user perception, such as direction, intensity, edge softness, and the handling of multiple light sources, to enable more targeted shadow-refinement mechanisms and multi-step adjustment strategies.

Additionally, near-real-time adjustments remain beyond our current scope. VLM inference is the primary bottleneck: image-based pipelines incur mean latencies of $4.43\text{ s}$ (placement) and $5.84\text{ s}$ (size), while video-based delays increase to $7.63\text{ s}$ and $9.62\text{ s}$, respectively. Future work will investigate fine-tuned edge VLMs to bypass latency-intensive cloud inference.

\paragraph{User Study Design} Our user study includes 21 participants and follows a fixed ordering in which participants evaluate the scene before and after VLM-guided adjustment, matching the natural system workflow. This design allows us to evaluate whether users perceive the adjusted result as better than the initial one, but it may also introduce expectation bias. In addition, some randomly initialized scenes already had high perceived quality, creating ceiling effects that reduce the observable room for improvement. In future work, we plan to use larger, diverse participant pools, systematically control scene quality initialization, and evaluate alternative study designs that better isolate adjustment effects.

\section{Conclusion}

In this paper, we use VLMs as scalable, human-aligned tools to assess and improve AR perceptual realism. We introduce \emph{RateAR}, a benchmark of $321$ AR images and $112$ AR videos captured from real AR renderings in physical environments, and use it to evaluate eleven commercial VLMs on three AR-critical visual features: placement plausibility, size appropriateness, and shadow realism. Our results show that VLM predictions closely track human subjective judgments, achieving SRCC up to $0.8695$. Building on these findings, we develop an autonomous AR content adjustment pipeline driven by VLM evaluations, and validate its utility in an IRB-approved user study with $21$ participants. Our study suggests VLMs can serve as perception-aware AR evaluators, providing effective feedback that reduces reliance on labor-intensive user studies and brittle geometry reconstruction.

\begin{acks}
We thank the participants of our user study for their invaluable help in this research and Tanish Pentakota for assistance implementing baselines. 

This work was supported in part by NSF grants CSR-2312760, CNS-2112562, and IIS-2231975, NSF CAREER Award IIS-2046072, NSF NAIAD Award 2332744, a CISCO Research Award, a Meta Research Award, Defense Advanced Research Projects Agency Young Faculty Award HR0011-24-1-0001, and the Army Research Laboratory under Cooperative Agreement Number W911NF-23-2-0224. The views and conclusions contained in this document are those of the authors. They should not be interpreted as representing the official policies, either expressed or implied, of the Defense Advanced Research Projects Agency, the Army Research Laboratory, or the U.S. Government. This paper has been approved for public release; distribution is unlimited. No official endorsement should be inferred. The U.S. Government is authorized to reproduce and distribute reprints for Government purposes, notwithstanding any copyright notation herein.
\end{acks}

\bibliographystyle{ACM-Reference-Format}
\bibliography{main}
\clearpage

\appendix
\setcounter{page}{1} 
\renewcommand{\thefigure}{A\arabic{figure}}
\setcounter{figure}{0} 
\renewcommand{\thetable}{A\arabic{table}}
\setcounter{table}{0} 
\section*{Supplemental Materials}
\addtocounter{section}{1} 
\nobalance
\subsection{VLM Content Adjustment Prompts} \label{sec:cont_adj_prompts}

We provide the VLM prompts used by the proposed autonomous content adjustment framework below.
\subsubsection{Placement Adjustment Task Prompt}

\leavevmode \\ 
\noindent
\begin{center}
    \begin{minipage}{0.9\linewidth} 
\begin{shaded}
\noindent\emph{Please assess whether the virtual object in this AR scene is correctly placed or appears to be floating/ misaligned. If it does float or look misaligned, indicate whether that behavior is appropriate for this object type. Rate placement appropriateness from 1 to 5, where 1 means an object that should be grounded/ aligned is clearly floating or misaligned, and 5 means a non-floating object is firmly grounded and properly aligned, or a float-capable object is reasonably floating. Use cues such as whether any rendered shadow (if present) is attached under the object, whether the object is intended for conceptual visualization, and whether the object interacts with forces or other objects (e.g., a ball bouncing under a hand). If the score is not 5, provide concrete, actionable adjustments in four directions (up, down, to the user’s left, and to the user’s right), listing only the directions that apply. Please provide me with 2 numbers to adjust the placement of the virtual content. The first number indicates the left-right adjustment, and the second number indicates the up-down adjustment. If it needs to move to the user’s left, the first number is negative, otherwise it is positive. If it needs to move upward, the second number is negative, otherwise it is positive. The unit of the number is meter. The range of the two numbers is between -1 and 1. If the virtual content is properly placed, please return 0 for both numbers. Please conclude with the 2 numbers in a json format like this: {`num1': -0.5, `num2': 0.5}.}
\end{shaded}
\end{minipage}
\end{center}

\subsubsection{Size Adjustment Task Prompt}
\leavevmode \\ 
\noindent
\begin{center}
    \begin{minipage}{0.9\linewidth} 
\begin{shaded}
\noindent\emph{Please assess whether the virtual object’s size appears normal for any plausible subtype of its category (e.g., a “table” could be a coffee table, console/sofa table, or desk) within this AR scene. If not, explain whether the atypical size is appropriate given the surrounding context. Rate size appropriateness on a 1–5 scale, where 1 means an object that should be normal-sized is clearly too large or too small, and 5 means the object is within a normal range or, if intentionally non-standard, is reasonably larger or smaller. Use cues such as nearby reference objects, human-scale affordances, and contextual clues (e.g., a surreal museum). If the score is not 5, provide concrete, actionable adjustments in two directions—Increase or Decrease—listing only those that apply. Please provide me with 1 number to adjust the scale of the virtual content. The number indicates the scale adjustment. If the number is greater than 1, it means to enlarge the virtual content, otherwise it means to shrink the virtual content. The range of the number is between 0.01 and 2. If the virtual content is properly scaled, please return 1. Please return the 1 number in a json format like this: {`num1': 1.0}.}
\end{shaded}
\end{minipage}
\end{center}

\subsection{Extended RateAR Baseline and VLM Performance Analysis} \label{sec:tables_CI_and_pvalues}

All correlation coefficient 95\% confidence intervals and statistical significance p-values are computed using the \texttt{Pingouin} package for Python~\cite{vallat2018pingouin}. \Cref{tab:main_results_CI} and \Cref{tab:main_results_pvals} report the confidence intervals and p-values corresponding to \Cref{tab:main_results} in the main text, respectively. Column-wise p-value adjustments are made using the Bonferroni correction method with $m_{images} = 18$ tests, such that $p_{adj}=\min(1,p_{raw} * m_{images})$. Similarly, \Cref{tab:video_results_CI} and \Cref{tab:video_results_pvals} present the matching auxiliary metrics for \Cref{tab:video_results}. RateAR video coefficient p-values are adjusted under the condition $m_{video}=17$.

\begin{table*}[hbtp]
\centering
\caption{Correlation metric 95\% confidence intervals between human scores and model predictions on RateAR images, corresponding to the coefficients in~\Cref {tab:main_results}. Confidence interval format: {[lower\_bound, upper\_bound]}.}
\label{tab:main_results_CI}
\resizebox{\textwidth}{!}{%
\begin{tabular}{@{}l|rrr|rrr|rrr}
\toprule
                             \multirow{2}{*}{\textbf{Methods}} & \multicolumn{3}{c|}{\begin{tabular}[c]{@{}c@{}}\textbf{Placement}\end{tabular}}                                    & \multicolumn{3}{c|}{\begin{tabular}[c]{@{}c@{}}\textbf{Size}\end{tabular}}                                      & \multicolumn{3}{c}{\begin{tabular}[c]{@{}c@{}}\textbf{Shadow}\end{tabular}}                                             \\ \cline{2-10} 
    & \multicolumn{1}{c}{SRCC}               & \multicolumn{1}{c}{PLCC}               & \multicolumn{1}{c|}{KRCC}              & \multicolumn{1}{c}{SRCC}               & \multicolumn{1}{c}{PLCC}               & \multicolumn{1}{c|}{KRCC}              & \multicolumn{1}{c}{SRCC}               & \multicolumn{1}{c}{PLCC}               & \multicolumn{1}{c}{KRCC}               \\ \hline
HyperIQA                   & 
{[0.40, 0.60]}                   & 
{[0.40, 0.60]}                   & 
{[0.25, 0.48]}                   & 
{[0.60, 0.74]}                   & 
{[0.56, 0.71]}                   & 
{[0.41, 0.61]}                   & 
{[0.40, 0.60]}                   & 
{[0.45, 0.63]}                   & 
{[0.24, 0.47]}                   \\
CNNIQA                     & 
{[0.24, 0.47]}                   & 
{[0.25, 0.48]}                   & 
{[0.13, 0.38]}                   & 
{[0.34, 0.55]}                   & 
{[0.36, 0.57]}                   & 
{[0.20, 0.44]}                   & 
{[0.19, 0.43]}                   & 
{[0.25, 0.48]}                   & 
{[0.09, 0.34]}                   \\
CLIPIQA+                   & 
{[0.26, 0.49]}                   & 
{[0.24, 0.47]}                   & 
{[0.15, 0.39]}                   & 
{[0.54, 0.70]}                   & 
{[0.48, 0.65]}                   & 
{[0.33, 0.54]}                   & 
{[0.29, 0.51]}                   & 
{[0.37, 0.57]}                   & 
{[0.16, 0.40]}                   \\
LIQE                       & 
{[0.35, 0.55]}                   & 
{[0.34, 0.55]}                   & 
{[0.21, 0.44]}                   & 
{[0.64, 0.77]}                   & 
{[0.61, 0.75]}                   & 
{[0.43, 0.62]}                   & 
{[0.39, 0.59]}                   & 
{[0.44, 0.63]}                   & 
{[0.23, 0.46]}                   \\
LPIPS-VGG+                & 
{[0.62, 0.76]}                   & 
{[0.57, 0.72]}                   & 
{[0.39, 0.59]}                   & 
{[0.47, 0.65]}                   & 
{[0.39, 0.59]}                   & 
{[0.27, 0.50]}                   & 
{[0.22, 0.46]}                   & 
{[0.21, 0.45]}                   & 
{[0.11, 0.36]}                   \\
TOPIQ                     & 
{[0.45, 0.64]}                   & 
{[0.41, 0.61]}                   & 
{[0.28, 0.50]}                   & 
{[0.59, 0.74]}                   & 
{[0.48, 0.66]}                   & 
{[0.38, 0.58]}                   & 
{[0.43, 0.62]}                   & 
{[0.47, 0.65]}                   & 
{[0.28, 0.50]}                   \\ 
QualiCLIP+                 & 
{[0.31, 0.52]}                   & 
{[0.34, 0.55]}                   & 
{[0.17, 0.41]}                   & 
{[0.62, 0.76]}                   & 
{[0.59, 0.73]}                   & 
{[0.42, 0.61]}                   & 
{[0.38, 0.58]}                   & 
{[0.46, 0.65]}                   & 
{[0.22, 0.45]}                   \\
\hline
Gemini-2.5-Flash           & 
{[0.50, 0.68]}                   & 
{[0.48, 0.66]}                   & 
{[0.37, 0.57]}                   & 
{[0.12, 0.37]}                   & 
{[0.17, 0.41]}                   & 
{[0.07, 0.32]}                   & 
{[0.34, 0.55]}                   & 
{[0.32, 0.53]}                   & 
{[0.21, 0.45]}                   \\
Gemini-2.5-Pro             & 
{[0.50, 0.67]}                   & 
{[0.47, 0.65]}                   & 
{[0.37, 0.58]}                   & 
{[0.21, 0.44]}                   & 
{[0.26, 0.48]}                   & 
{[0.14, 0.39]}                   & 
{[0.27, 0.49]}                   & 
{[0.18, 0.42]}                   & 
{[0.18, 0.42]}                   \\
Gemini-3.5-Flash           & 
{[0.67, 0.79]}                   & 
{[0.68, 0.80]}                   & 
{[0.52, 0.68]}                   & 
{[0.57, 0.73]}                   & 
{[0.66, 0.78]}                   & 
{[0.43, 0.62]}                   & 
{[0.37, 0.58]}                   & 
{[0.33, 0.54]}                   & 
{[0.27, 0.49]}                   \\
Gemini-3.1-Pro             & 
{[0.37, 0.57]}                   & 
{[0.31, 0.53]}                   & 
{[0.27, 0.50]}                   & 
{[0.36, 0.57]}                   & 
{[0.40, 0.60]}                   & 
{[0.25, 0.48]}                   & 
{[0.18, 0.42]}                   & 
{[0.16, 0.40]}                   & 
{[0.12, 0.37]}                   \\
GPT-o3                     & 
{[0.68, 0.80]}                   & 
{[0.64, 0.77]}                   & 
{[0.52, 0.68]}                   & 
{[0.67, 0.79]}                   & 
{[0.69, 0.81]}                   & 
{[0.49, 0.66]}                   & 
{[0.49, 0.67]}                   & 
{[0.44, 0.63]}                   & 
{[0.36, 0.56]}                   \\
GPT-5                      & 
{[0.79, 0.87]}                   & 
{[0.74, 0.84]}                   & 
{[0.61, 0.75]}                   & 
{[0.39, 0.59]}                   & 
{[0.49, 0.66]}                   & 
{[0.27, 0.49]}                   & 
{[0.40, 0.60]}                   & 
{[0.39, 0.59]}                   & 
{[0.28, 0.50]}                   \\
GPT-5.4-Mini               & 
{[0.47, 0.65]}                   & 
{[0.45, 0.64]}                   & 
{[0.35, 0.56]}                   & 
{[-0.09, 0.17]}                   & 
{[-0.04, 0.22]}                   & 
{[-0.10, 0.17]}                   & 
{[0.22, 0.45]}                   & 
{[0.20, 0.43]}                   & 
{[0.14, 0.38]}                   \\
GPT-5.4                    & 
{[0.65, 0.78]}                   & 
{[0.62, 0.76]}                   & 
{[0.49, 0.67]}                   & 
{[0.08, 0.33]}                   & 
{[0.22, 0.46]}                   & 
{[0.03, 0.28]}                   & 
{[0.52, 0.68]}                   & 
{[0.49, 0.66]}                   & 
{[0.38, 0.58]}                   \\
Grok-4.3                   & 
{[0.49, 0.67]}                   & 
{[0.47, 0.65]}                   & 
{[0.37, 0.57]}                   & 
{[0.04, 0.30]}                   & 
{[0.10, 0.35]}                   & 
{[0.00, 0.26]}                   & 
{[0.49, 0.67]}                   & 
{[0.49, 0.67]}                   & 
{[0.34, 0.55]}                   \\
Grok-4.5                   & 
{[0.52, 0.69]}                   & 
{[0.52, 0.69]}                   & 
{[0.38, 0.58]}                   & 
{[0.38, 0.58]}                   & 
{[0.47, 0.65]}                   & 
{[0.27, 0.50]}                   & 
{[0.51, 0.68]}                   & 
{[0.48, 0.66]}                   & 
{[0.34, 0.55]}                   \\
Muse-Spark-1.1             & 
{[0.73, 0.83]}                   & 
{[0.73, 0.83]}                   & 
{[0.56, 0.72]}                   & 
{[0.59, 0.74]}                   & 
{[0.62, 0.76]}                   & 
{[0.44, 0.62]}                   & 
{[0.35, 0.55]}                   & 
{[0.36, 0.57]}                   & 
{[0.24, 0.47]}                   \\
\bottomrule
\end{tabular}}
\end{table*}

\begin{table*}[hbtp] 
\centering
\caption{Correlation coefficient statistical significance p-values between human scores and model predictions on RateAR images, corresponding to the coefficients in~\Cref {tab:main_results}. P-Value format: {$p_{adj} \ (p_{raw})$; for highly significant results (i.e., $p_{adj}<.001$), only the adjusted p-value is reported.}}
\label{tab:main_results_pvals}
\begin{tabular}{@{}l|ccc|ccc|ccc}
\toprule
                             \multirow{2}{*}{\textbf{Methods}} & \multicolumn{3}{c|}{\begin{tabular}[c]{@{}c@{}}\textbf{Placement}\end{tabular}}                                    & \multicolumn{3}{c|}{\begin{tabular}[c]{@{}c@{}}\textbf{Size}\end{tabular}}                                      & \multicolumn{3}{c}{\begin{tabular}[c]{@{}c@{}}\textbf{Shadow}\end{tabular}}                                             \\ \cline{2-10} 
    & \multicolumn{1}{c}{SRCC}               & \multicolumn{1}{c}{PLCC}               & \multicolumn{1}{c|}{KRCC}              & \multicolumn{1}{c}{SRCC}               & \multicolumn{1}{c}{PLCC}               & \multicolumn{1}{c|}{KRCC}              & \multicolumn{1}{c}{SRCC}               & \multicolumn{1}{c}{PLCC}               & \multicolumn{1}{c}{KRCC}               \\ \hline
HyperIQA                   & 
{<.001}                   & 
{<.001}                   &
{<.001}                   &
{<.001}                   &
{<.001}                   &
{<.001}                   &
{<.001}                   &
{<.001}                   &
{<.001}                   \\
CNNIQA                     & 
{<.001}                   & 
{<.001}                   &
{<.001}                   &
{<.001}                   &
{<.001}                   &
{<.001}                   &
{<.001}                   &
{<.001}                   &
{<.001}                   \\
CLIPIQA+                   & 
{<.001}                   & 
{<.001}                   &
{<.001}                   &
{<.001}                   &
{<.001}                   &
{<.001}                   &
{<.001}                   &
{<.001}                   &
{<.001}                   \\
LIQE                       & 
{<.001}                   & 
{<.001}                   &
{<.001}                   &
{<.001}                   &
{<.001}                   &
{<.001}                   &
{<.001}                   &
{<.001}                   &
{<.001}                   \\
LPIPS-VGG+                & 
{<.001}                   & 
{<.001}                   &
{<.001}                   &
{<.001}                   &
{<.001}                   &
{<.001}                   &
{<.001}                   &
{<.001}                   &
{<.001}                   \\
TOPIQ                     & 
{<.001}                   & 
{<.001}                   &
{<.001}                   &
{<.001}                   &
{<.001}                   &
{<.001}                   &
{<.001}                   &
{<.001}                   &
{<.001}                   \\ 
QualiCLIP+                 & 
{<.001}                   & 
{<.001}                   &
{<.001}                   &
{<.001}                   &
{<.001}                   &
{<.001}                   &
{<.001}                   &
{<.001}                   &
{<.001}                   \\
\hline
Gemini-2.5-Flash           & 
{<.001}                   & 
{<.001}                   & 
{<.001}                   & 
{.004 (<.001)}            & 
{<.001}                   & 
{.004 (<.001)}            & 
{<.001}                   &
{<.001}                   &
{<.001}                   \\
Gemini-2.5-Pro             & 
{<.001}                   & 
{<.001}                   &
{<.001}                   &
{<.001}                   &
{<.001}                   &
{<.001}                   &
{<.001}                   &
{<.001}                   &
{<.001}                   \\
Gemini-3.5-Flash           & 
{<.001}                   & 
{<.001}                   &
{<.001}                   &
{<.001}                   &
{<.001}                   &
{<.001}                   &
{<.001}                   &
{<.001}                   &
{<.001}                   \\
Gemini-3.1-Pro             & 
{<.001}                   & 
{<.001}                   &
{<.001}                   &
{<.001}                   &
{<.001}                   &
{<.001}                   &
{<.001}                   &
{<.001}                   &
{<.001}                   \\
GPT-o3                     & 
{<.001}                   & 
{<.001}                   &
{<.001}                   &
{<.001}                   &
{<.001}                   &
{<.001}                   &
{<.001}                   &
{<.001}                   &
{<.001}                   \\
GPT-5                      & 
{<.001}                   & 
{<.001}                   &
{<.001}                   &
{<.001}                   &
{<.001}                   &
{<.001}                   &
{<.001}                   &
{<.001}                   &
{<.001}                   \\
GPT-5.4-Mini               & 
{<.001}                   & 
{<.001}                   &
{<.001}                   & 
{>.999 (.519)}            & 
{>.999 (.189)}            & 
{>.999 (.517)}            & 
{<.001}                   &
{<.001}                   &
{<.001}                   \\
GPT-5.4                    & 
{<.001}                   & 
{<.001}                   &
{<.001}                   & 
{.029 (.002)}             & 
{<.001}                   & 
{.044 (.002)}             & 
{<.001}                   &
{<.001}                   &
{<.001}                   \\
Grok-4.3                   & 
{<.001}                   & 
{<.001}                   &
{<.001}                   & 
{.199 (.011)}             & 
{.013 (<.001)}            & 
{.179 (.010)}             & 
{<.001}                   &
{<.001}                   &
{<.001}                   \\
Grok-4.5                   & 
{<.001}                   & 
{<.001}                   &
{<.001}                   &
{<.001}                   &
{<.001}                   &
{<.001}                   &
{<.001}                   &
{<.001}                   &
{<.001}                   \\
Muse-Spark-1.1             & 
{<.001}                   & 
{<.001}                   &
{<.001}                   &
{<.001}                   &
{<.001}                   &
{<.001}                   &
{<.001}                   &
{<.001}                   &
{<.001}                   \\
\bottomrule
\end{tabular}
\end{table*}

\begin{table*}[hbtp] 
\centering
\caption{Correlation metric 95\% confidence intervals between human scores and model predictions on RateAR videos, corresponding to the coefficients in~\Cref {tab:video_results}. Confidence interval format: {[lower\_bound, upper\_bound]}.}
\label{tab:video_results_CI}
\resizebox{\textwidth}{!}{%
\begin{tabular}{@{}l|rrr|rrr|rrr}
\toprule
                             \multirow{2}{*}{\textbf{Methods}} & \multicolumn{3}{c|}{\begin{tabular}[c]{@{}c@{}}\textbf{Placement}\end{tabular}}                                    & \multicolumn{3}{c|}{\begin{tabular}[c]{@{}c@{}}\textbf{Size}\end{tabular}}                                      & \multicolumn{3}{c}{\begin{tabular}[c]{@{}c@{}}\textbf{Shadow}\end{tabular}}                                             \\ \cline{2-10} 
    & \multicolumn{1}{c}{SRCC}               & \multicolumn{1}{c}{PLCC}               & \multicolumn{1}{c|}{KRCC}              & \multicolumn{1}{c}{SRCC}               & \multicolumn{1}{c}{PLCC}               & \multicolumn{1}{c|}{KRCC}              & \multicolumn{1}{c}{SRCC}               & \multicolumn{1}{c}{PLCC}               & \multicolumn{1}{c}{KRCC}               \\ \hline
DOVER                      & 
{[0.22, 0.62]}                   & 
{[0.21, 0.62]}                   & 
{[0.08, 0.53]}                   & 
{[0.30, 0.67]}                   & 
{[0.44, 0.75]}                   & 
{[0.12, 0.56]}                   & 
{[-0.26, 0.23]}                   & 
{[-0.19, 0.31]}                   & 
{[-0.26, 0.24]}                   \\
COVER                      & 
{[0.14, 0.58]}                   & 
{[0.19, 0.61]}                   & 
{[0.02, 0.49]}                   & 
{[0.24, 0.64]}                   & 
{[0.40, 0.73]}                   & 
{[0.07, 0.52]}                   & 
{[-0.21, 0.29]}                   & 
{[-0.29, 0.21]}                   & 
{[-0.21, 0.28]}                   \\ ModularBVQA                & 
{[0.21, 0.62]}                   & 
{[0.21, 0.62]}                   & 
{[0.07, 0.52]}                   & 
{[0.46, 0.76]}                   & 
{[0.65, 0.86]}                   & 
{[0.22, 0.63]}                   & 
{[-0.03, 0.44]}                   & 
{[-0.11, 0.38]}                   & 
{[-0.11, 0.38]}                   \\
ReLaX-VQA                  & 
{[0.10, 0.54]}                   & 
{[0.10, 0.55]}                   & 
{[-0.04, 0.44]}                   & 
{[0.26, 0.65]}                   & 
{[0.39, 0.72]}                   & 
{[0.11, 0.55]}                   & 
{[-0.28, 0.21]}                   & 
{[-0.34, 0.15]}                   & 
{[-0.27, 0.23]}                   \\
E-VQA                      & 
{[-0.17, 0.33]}                   & 
{[-0.16, 0.34]}                   & 
{[-0.20, 0.30]}                   & 
{[0.03, 0.49]}                   & 
{[0.10, 0.54]}                   & 
{[-0.06, 0.42]}                   & 
{[-0.23, 0.26]}                   & 
{[-0.17, 0.33]}                   & 
{[-0.24, 0.26]}                   \\
DIVA-VQA                   & 
{[0.09, 0.53]}                   & 
{[0.13, 0.56]}                   & 
{[-0.01, 0.46]}                   & 
{[0.29, 0.67]}                   & 
{[0.45, 0.76]}                   & 
{[0.14, 0.57]}                   & 
{[-0.15, 0.34]}                   & 
{[-0.18, 0.32]}                   & 
{[-0.19, 0.31]}                   \\
\hline
Gemini-2.5-Flash           & 
{[0.52, 0.79]}                   & 
{[0.51, 0.79]}                   & 
{[0.38, 0.72]}                   & 
{[0.01, 0.48]}                   & 
{[0.07, 0.52]}                   & 
{[-0.05, 0.43]}                   & 
{[0.11, 0.55]}                   & 
{[0.16, 0.59]}                   & 
{[0.03, 0.50]}                   \\
Gemini-2.5-Pro             & 
{[0.23, 0.63]}                   & 
{[0.22, 0.63]}                   & 
{[0.13, 0.57]}                   & 
{[0.10, 0.54]}                   & 
{[0.19, 0.61]}                   & 
{[0.04, 0.50]}                   & 
{[0.27, 0.66]}                   & 
{[0.17, 0.60]}                   & 
{[0.15, 0.58]}                   \\
Gemini-3.5-Flash           & 
{[0.56, 0.81]}                   & 
{[0.59, 0.83]}                   & 
{[0.44, 0.75]}                   & 
{[0.11, 0.56]}                   & 
{[0.23, 0.63]}                   & 
{[0.05, 0.51]}                   & 
{[0.22, 0.63]}                   & 
{[0.11, 0.55]}                   & 
{[0.13, 0.57]}                   \\
Gemini-3.1-Pro             & 
{[0.17, 0.59]}                   & 
{[0.19, 0.61]}                   & 
{[0.11, 0.55]}                   & 
{[-0.07, 0.42]}                   & 
{[-0.03, 0.44]}                   & 
{[-0.10, 0.39]}                   & 
{[-0.06, 0.42]}                   & 
{[-0.16, 0.33]}                   & 
{[-0.09, 0.39]}                   \\
GPT-o3                     & 
{[0.56, 0.81]}                   & 
{[0.65, 0.86]}                   & 
{[0.39, 0.73]}                   & 
{[0.45, 0.76]}                   & 
{[0.55, 0.81]}                   & 
{[0.31, 0.68]}                   & 
{[0.44, 0.76]}                   & 
{[0.41, 0.74]}                   & 
{[0.29, 0.67]}                   \\
GPT-5                     & 
{[0.76, 0.90]}                   & 
{[0.66, 0.86]}                   & 
{[0.55, 0.81]}                   & 
{[0.72, 0.89]}                   & 
{[0.72, 0.89]}                   & 
{[0.53, 0.80]}                   & 
{[0.79, 0.92]}                   & 
{[0.75, 0.90]}                   & 
{[0.62, 0.84]}                   \\
GPT-5.4-Mini               & 
{[0.25, 0.64]}                   & 
{[0.21, 0.62]}                   & 
{[0.15, 0.58]}                   & 
{[0.24, 0.64]}                   & 
{[0.23, 0.63]}                   & 
{[0.16, 0.59]}                   & 
{[0.40, 0.73]}                   & 
{[0.51, 0.79]}                   & 
{[0.29, 0.67]}                   \\
GPT-5.4                    & 
{[0.29, 0.67]}                   & 
{[0.37, 0.72]}                   & 
{[0.18, 0.60]}                   & 
{[0.22, 0.63]}                   & 
{[0.22, 0.62]}                   & 
{[0.13, 0.57]}                   & 
{[0.71, 0.89]}                   & 
{[0.75, 0.90]}                   & 
{[0.56, 0.81]}                   \\
Grok-4.3                   & 
{[0.47, 0.77]}                   & 
{[0.46, 0.76]}                   & 
{[0.35, 0.71]}                   & 
{[0.38, 0.72]}                   & 
{[0.44, 0.75]}                   & 
{[0.27, 0.66]}                  & 
{[0.72, 0.89]}                   & 
{[0.84, 0.94]}                   & 
{[0.51, 0.79]}                   \\
Grok-4.5                   & 
{[0.65, 0.86]}                   & 
{[0.55, 0.81]}                   & 
{[0.48, 0.77]}                   & 
{[0.17, 0.59]}                   & 
{[0.23, 0.63]}                   & 
{[0.09, 0.54]}                  & 
{[0.61, 0.84]}                   & 
{[0.68, 0.87]}                   & 
{[0.46, 0.76]}                   \\
Muse-Spark-1.1            & 
{[0.69, 0.88]}                   & 
{[0.68, 0.87]}                   & 
{[0.53, 0.80]}                   & 
{[0.29, 0.67]}                   & 
{[0.37, 0.71]}                   & 
{[0.18, 0.60]}                  & 
{[0.23, 0.63]}                   & 
{[0.27, 0.66]}                   & 
{[0.15, 0.58]}                   \\ \bottomrule
\end{tabular}}
\end{table*}

\begin{table*}[hbtp]
\centering
\caption{Correlation coefficient statistical significance p-values between human scores and model predictions on RateAR videos, corresponding to the coefficients in~\Cref {tab:video_results}. P-Value format: {$p_{adj} \ (p_{raw})$; for highly significant results (i.e., $p_{adj}<.001$), only the adjusted p-value is reported}.}
\label{tab:video_results_pvals}
\resizebox{\textwidth}{!}{%
\begin{tabular}{@{}l|ccc|ccc|ccc}
\toprule
                             \multirow{2}{*}{\textbf{Methods}} & \multicolumn{3}{c|}{\begin{tabular}[c]{@{}c@{}}\textbf{Placement}\end{tabular}}                                    & \multicolumn{3}{c|}{\begin{tabular}[c]{@{}c@{}}\textbf{Size}\end{tabular}}                                      & \multicolumn{3}{c}{\begin{tabular}[c]{@{}c@{}}\textbf{Shadow}\end{tabular}}                                             \\ \cline{2-10} 
    & \multicolumn{1}{c}{SRCC}               & \multicolumn{1}{c}{PLCC}               & \multicolumn{1}{c|}{KRCC}              & \multicolumn{1}{c}{SRCC}               & \multicolumn{1}{c}{PLCC}               & \multicolumn{1}{c|}{KRCC}              & \multicolumn{1}{c}{SRCC}               & \multicolumn{1}{c}{PLCC}               & \multicolumn{1}{c}{KRCC}               \\ \hline
DOVER                      & 
{.005 (<.001)}                   & 
{.006 (<.001)}                   & 
{.009 (<.001)}                   & 
{<.001}                   & 
{<.001}                   & 
{.001 (<.001)}                   & 
{>.999 (.797)}                   & 
{>.999 (.635)}                   & 
{>.999 (.815)}                   \\
COVER                      & 
{.039 (.002)}                   & 
{.010 (<.001)}                   & 
{.059 (.003)}                   & 
{.003 (<.001)}                   & 
{<.001}                   & 
{.009 (<.001)}                   & 
{>.999 (.728)}                   & 
{>.999 (.767)}                   & 
{>.999 (.685)}                   \\ 
ModularBVQA                & 
{.007 (<.001)}                   & 
{.006 (<.001)}                   & 
{.012 (<.001)}                   & 
{<.001}                   & 
{<.001}                   & 
{<.001}                   & 
{>.999 (.087)}                   & 
{>.999 (.265)}                   & 
{>.999 (.110)}                   \\
ReLaX-VQA                  & 
{.121 (.007)}                   & 
{.101 (.006)}                   & 
{.352 (.021)}                   & 
{.002 (<.001)}                   & 
{<.001}                   & 
{.002 (<.001)}                   & 
{>.999 (.775)}                   & 
{>.999 (.427)}                   & 
{>.999 (.815)}                   \\
E-VQA                      & 
{>.999 (.521)}                   & 
{>.999 (.469)}                   & 
{>.999 (.542)}                   & 
{.477 (.028)}                   & 
{.113 (.007)}                   & 
{.604 (.036)}                   & 
{>.999 (.901)}                   & 
{>.999 (.507)}                   & 
{>.999 (.902)}                   \\
DIVA-VQA                   & 
{.153 (.009)}                   & 
{.057 (.003)}                   & 
{.153 (.009)}                   & 
{<.001}                   & 
{<.001}                   & 
{<.001}                   & 
{>.999 (.443)}                   & 
{>.999 (.561)}                   & 
{>.999 (.468)}                   \\
\hline
Gemini-2.5-Flash           & 
{<.001}                   & 
{<.001}                   & 
{<.001}                   & 
{.749 (.044)}                   & 
{.237 (.014)}                   & 
{.767 (.045)}                   & 
{.082 (.005)}                   & 
{.026 (.002)}                   & 
{.082 (.005)}                   \\
Gemini-2.5-Pro             & 
{.003 (<.001)}                   & 
{.005 (<.001)}                   & 
{.011 (<.001)}                   & 
{.116 (.007)}                   & 
{.011 (<.001)}                   & 
{.130 (.008)}                   & 
{.001 (<.001)}                   & 
{.018 (.001)}                   & 
{.002 (<.001)}                   \\
Gemini-3.5-Flash           & 
{<.001}                   & 
{<.001}                   & 
{<.001}                   & 
{.080 (.005)}                   & 
{.004 (<.001)}                   & 
{.085 (.005)}                   & 
{.004 (<.001)}                   & 
{.090 (.005)}                   & 
{.004 (<.001)}                   \\
Gemini-3.1-Pro             & 
{.020 (.001)}                   & 
{.012 (<.001)}                   & 
{.028 (.002)}                   & 
{>.999 (.147)}                   & 
{>.999 (.087)}                   & 
{>.999 (.148)}                   & 
{>.999 (.143)}                   & 
{>.999 (.484)}                   & 
{>.999 (.139)}                   \\
GPT-o3                     & 
{<.001}                   & 
{<.001}                   &
{<.001}                   &
{<.001}                   &
{<.001}                   &
{<.001}                   &
{<.001}                   &
{<.001}                   &
{<.001}                   \\
GPT-5                     & 
{<.001}                   & 
{<.001}                   &
{<.001}                   &
{<.001}                   &
{<.001}                   &
{<.001}                   &
{<.001}                   &
{<.001}                   &
{<.001}                   \\
GPT-5.4-Mini               & 
{.002 (<.001)}                   & 
{.006 (<.001)}                   & 
{.003 (<.001)}                   & 
{.003 (<.001)}                   & 
{.004 (<.001)}                   & 
{.005 (<.001)}                   & 
{<.001}                   & 
{<.001}                   & 
{<.001}                   \\
GPT-5.4                    & 
{<.001}                   & 
{<.001}                   & 
{.001 (<.001)}                   & 
{.005 (<.001)}                   & 
{.005 (<.001)}                   & 
{.007 (<.001)}                   & 
{<.001}                   & 
{<.001}                   & 
{<.001}                   \\
Grok-4.3                   & 
{<.001}                   & 
{<.001}                   &
{<.001}                   &
{<.001}                   &
{<.001}                   &
{<.001}                   &
{<.001}                   &
{<.001}                   &
{<.001}                   \\
Grok-4.5                   & 
{<.001}                   & 
{<.001}                   & 
{<.001}                   & 
{.022 (.001)}                   & 
{.003 (<.001)}                   & 
{.032 (.002)}                  & 
{<.001}                   & 
{<.001}                   & 
{<.001}                   \\
Muse-Spark-1.1            & 
{<.001}                   & 
{<.001}                   & 
{<.001}                   & 
{<.001}                   & 
{<.001}                   & 
{.001 (<.001)}                  & 
{.003 (<.001)}                   & 
{.001 (<.001)}                   & 
{.002 (<.001)}                   \\ \bottomrule
\end{tabular}}
\end{table*}

\subsection{Extended User Study Analysis} \label{sec:extend_user_study}

\begin{figure*}[hbtp]
\centering
\includegraphics[width=0.97\textwidth]{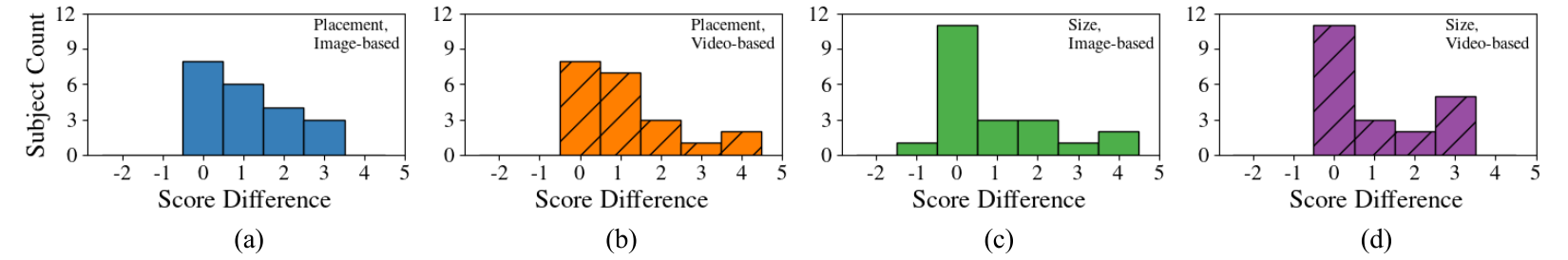}
\caption[Pre–post changes in human-rated quality for placement and size. General trends reflect left-skewed patterns toward zero, but more users report feature improvements than not.]{Pre–post changes in human-rated quality for placement, via (a) image-based and (b) video-based pipelines, and size, via (c) image-based and (d) video-based pipelines. Scores indicate improvement (positive), no change (zero), or degradation (negative) after single-iteration VLM adjustment.}
\Description{Pre–post changes in human-rated quality for placement and size. General trends reflect left-skewed patterns toward zero, but more users report feature improvements than not.}
\label{fig:score_change}
\end{figure*}

\Cref{fig:score_change} presents the differences in pre-post adjustment scores recorded by user study participants, as first introduced in~\Cref{sec:user_study_design,sec:user_study_results}. Furthermore, the approximate p-values for the conducted paired Wilcoxon signed-rank tests are reported in~\Cref{stest_pval}. As shown in the table, all p-values are below $.05$, and all p-values except that of the image-based size test are below $.01$.

\begin{table*} [hbtp]
    \centering
    \caption{Paired Wilcoxon signed-rank test p-values for statistical testing reported in~\Cref{sec:adj_performance}}
    \begin{tabular}{l|r} \toprule
        Trial & p-Value \\ \hline
        Placement, Image-based & 0.0012 \\
        Placement, Video-based & 0.0012 \\
        Size, Image-based & 0.0101 \\
        Size, Video-based & 0.0044 \\ \bottomrule
    \end{tabular}
    \label{stest_pval}
\end{table*}

\subsection{Ablation Study: Primitive and Few-Shot Prompts}

We provide the VLM prompts constructed for the ablation experiments. Recall that the Contextual prompts refer to the same text presented in~\Cref{sec:Prompt Template Design}. Hence, we only provide the primitive and few-shot prompts below. 

\subsubsection{Primitive Prompts} \label{sec:ablation_primative}

\paragraph{Placement Plausibility}
\leavevmode \\ 
\noindent
\begin{center}
    \begin{minipage}{0.9\linewidth} 
\begin{shaded}
\noindent\emph{Please assess whether the virtual object in this AR scene is correctly placed or appears to be floating/misaligned. Rate placement appropriateness from 1 to 5, and conclude the assessment with a final sentence in this exact format: Score: [your rating].}
\end{shaded}
\end{minipage}
\end{center}

\paragraph{Size Appropriateness}
\leavevmode \\ 
\noindent
\begin{center}
    \begin{minipage}{0.9\linewidth} 
\begin{shaded}
\noindent\emph{Please assess whether the virtual object’s size appears normal/reasonable within this AR scene. Rate size appropriateness on a 1–5 scale, and conclude the assessment with a final sentence in this exact format: Score: [your rating].}
\end{shaded}
\end{minipage}
\end{center}

\paragraph{Shadow Realism}
\leavevmode \\ 
\noindent
\begin{center}
    \begin{minipage}{0.9\linewidth} 
\begin{shaded}
\noindent\emph{Please assess whether the virtual object’s shadow is appropriate for this AR scene. Rate shadow appropriateness on a 1–5 scale, and conclude the assessment with a final sentence in this exact format: Score: [your rating].}
\end{shaded}
\end{minipage}
\end{center}

\clearpage
\balance

\subsubsection{Few-Shot Prompts} \label{sec:ablation_few_shot}

\paragraph{Placement Plausibility}

\leavevmode \\ 
\noindent
\begin{center}
    \begin{minipage}{0.9\linewidth} 
\begin{shaded}
\noindent\emph{Task: ``You are an expert in observing, describing, and estimating augmented reality images. You 
are tasked with rating the placement plausibility of augmented reality (AR) content. Please assess whether the virtual object in the AR scene is correctly placed or appears to 
be floating/misaligned. If it does float or look misaligned, indicate whether that behavior is appropriate for this object type. Rate placement appropriateness from 1 to 5, where 1 means an object that should be grounded/aligned is clearly floating or misaligned, and 5 means a non-floating object is firmly grounded and properly aligned, or a float-capable object is reasonably floating. Use cues such as whether any rendered shadow (if present) is attached under the object, whether the object is intended for conceptual visualization, and whether the object interacts with forces or other objects (e.g., a ball bouncing under a hand). Please end with a final sentence in this exact format: Score: [your rating]. You are provided with three AR scene examples and their perceived placement plausibility scores.''
}

\noindent\emph{Input:}
\end{shaded}
\end{minipage}
\end{center}
\begin{figure}[H]
    \centering
    \includegraphics[width=0.35\linewidth]{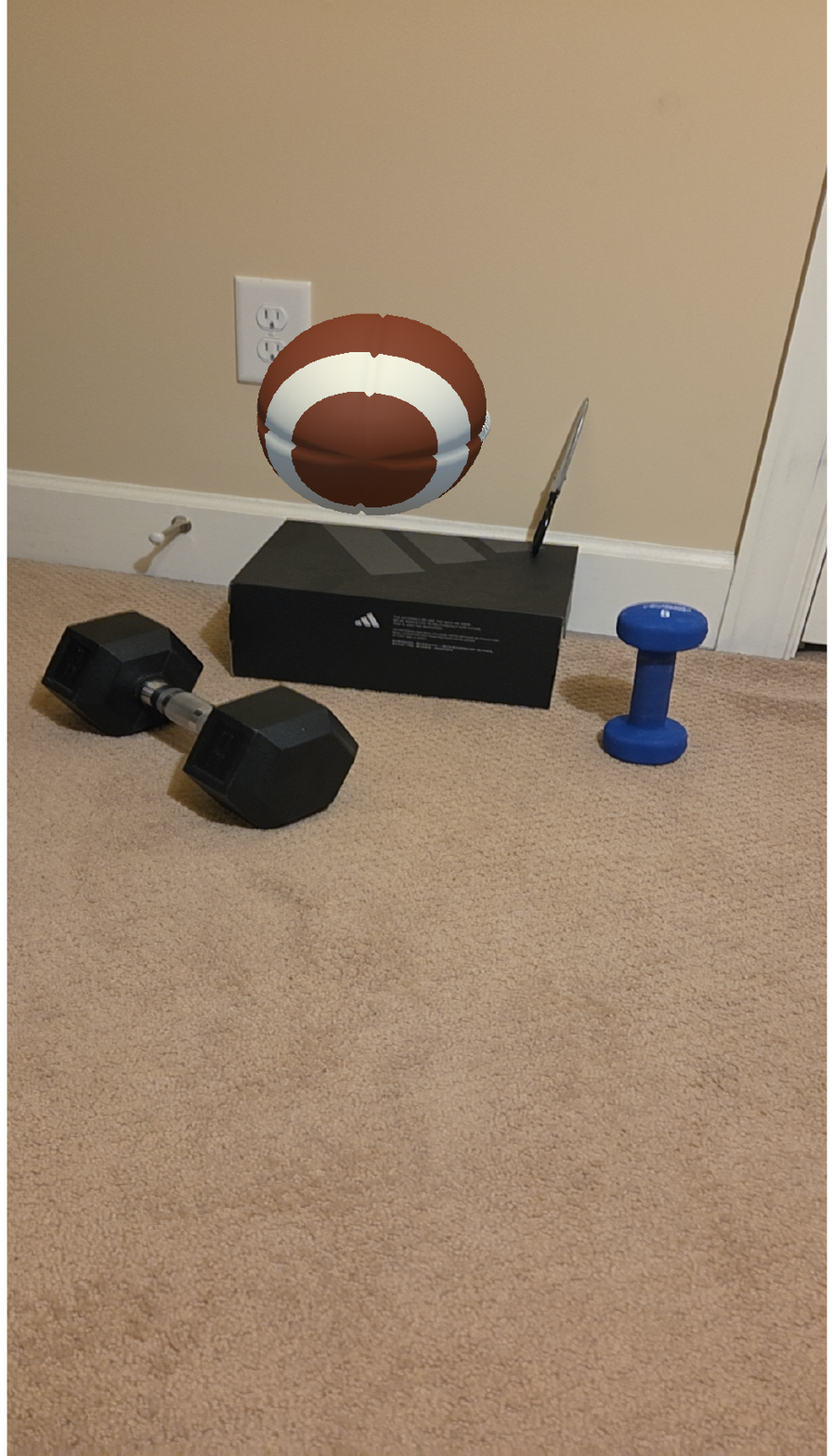}
    \label{fig:fewshot_place_1}
    \Description{Few-shot prompting instance with an assigned placement plausibility score of 1.}
\end{figure}

\noindent
\begin{center}
    \begin{minipage}{0.9\linewidth} 
\begin{shaded}
\noindent\emph{Output: ``The augmented reality content is a football. Score: 1''}

\noindent\emph{Input:}
\end{shaded}
\end{minipage}
\end{center}

\begin{figure}[H]
    \centering
    \includegraphics[width=0.35\linewidth]{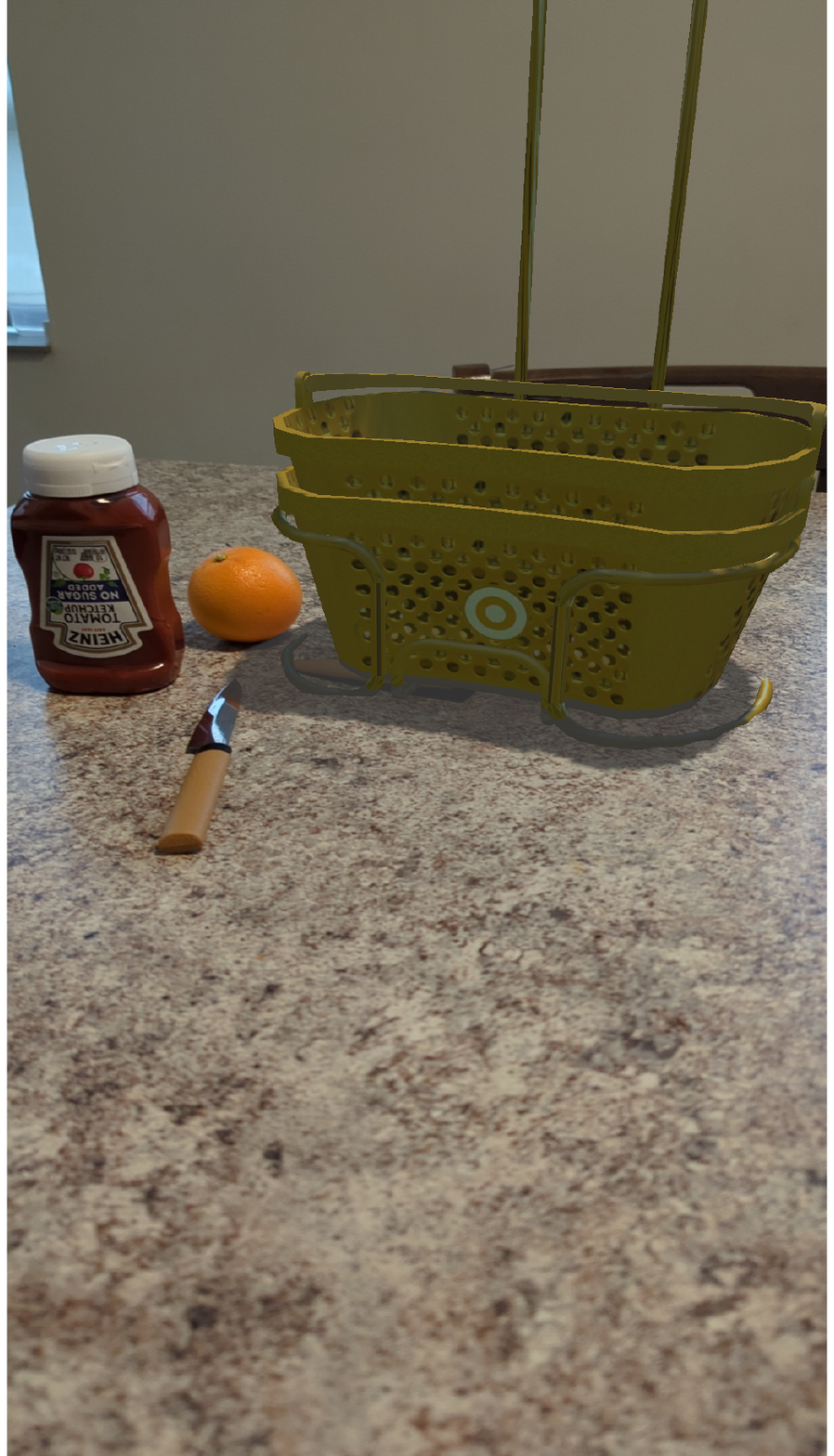}
    \label{fig:fewshot_place_5}
    \Description{Few-shot prompting instance with an assigned placement plausibility score of 5.}
\end{figure}

\noindent
\begin{center}
    \begin{minipage}{0.9\linewidth} 
\begin{shaded}
\noindent\emph{Output: ``The augmented reality content is a basket. Score: 5''}

\noindent\emph{Input:}
\end{shaded}
\end{minipage}
\end{center}

\begin{figure}[H]
    \centering
    \includegraphics[width=0.35\linewidth]{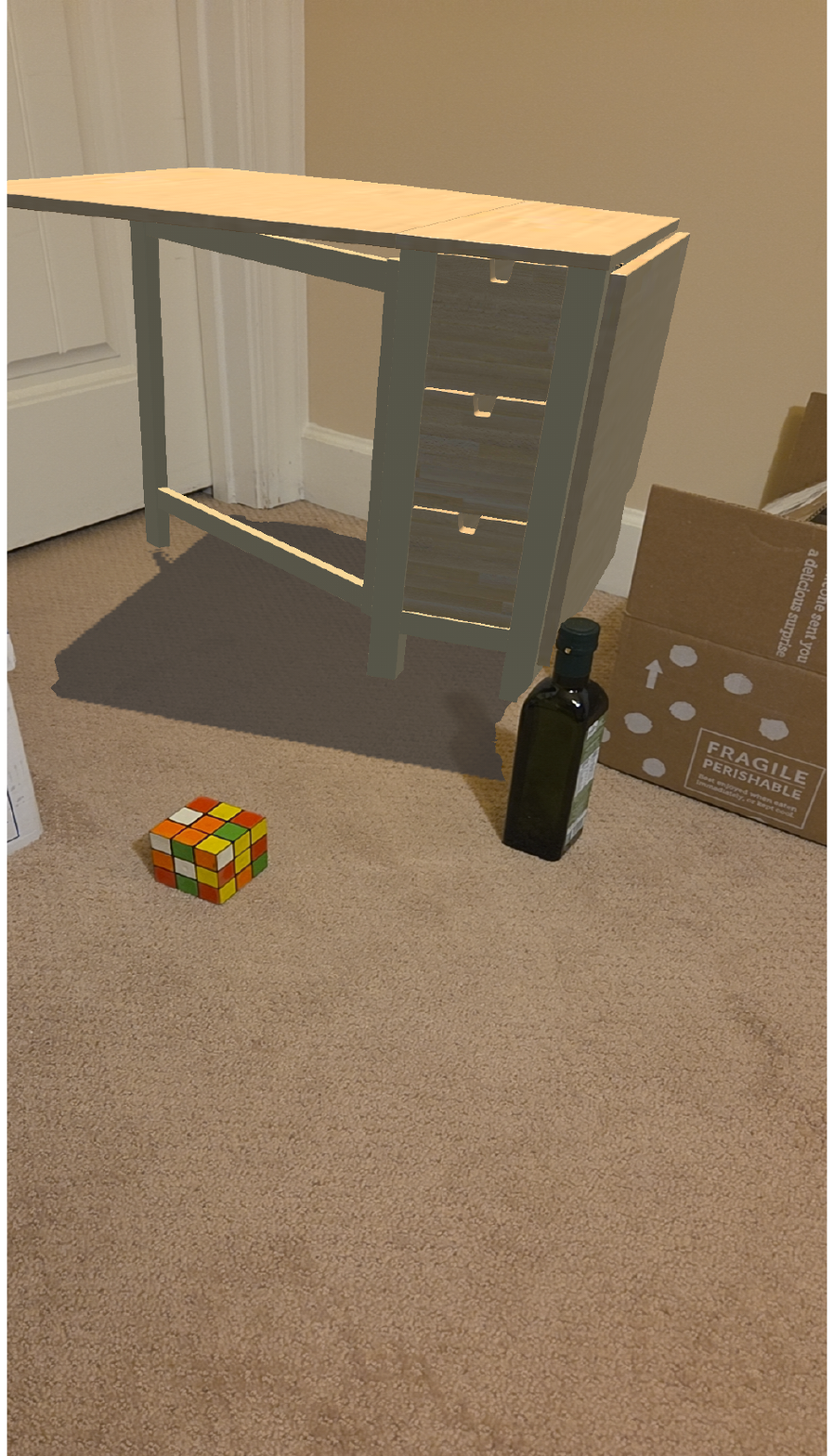}
    \label{fig:fewshot_place_3}
    \Description{Few-shot prompting instance with an assigned placement plausibility score of 3.}
\end{figure}
\noindent
\begin{center}
    \begin{minipage}{0.9\linewidth} 
\begin{shaded}
\noindent\emph{Output: ``The augmented reality content is a wooden desk-table. Score: 3''}
\newline
\noindent\emph{Input: [Target image to analyze]}
\end{shaded}
\end{minipage}
\end{center}

\paragraph{Size Appropriateness}
\leavevmode \\ 
\noindent
\begin{center}
    \begin{minipage}{0.9\linewidth} 
\begin{shaded}
\noindent Task: \emph{``You are an expert in observing, describing, and estimating augmented reality images. You are tasked with rating the size appropriateness of augmented reality (AR) content. Please assess whether the virtual object’s size appears normal for any plausible subtype of its category (e.g., a “table” could be a coffee table, console/sofa table, or desk) within the AR scene. If not, explain whether the atypical size is appropriate given the surrounding context. Rate size appropriateness on a 1–5 scale, where 1 means an object that should be normal-sized is clearly too large or too small, and 5 means the object is within a normal range or, if intentionally non-standard, is reasonably larger or smaller. Use cues such as nearby reference objects, human-scale affordances, and contextual clues (e.g., a surreal museum). Please end with a final sentence in this exact format: Score: [your rating]. You are provided with three AR scene examples and their perceived size appropriateness scores.''}
\newline
\noindent\emph{Input:}
\end{shaded}
\end{minipage}
\end{center}
\begin{figure}[H]
    \centering
    \includegraphics[width=0.5\linewidth]{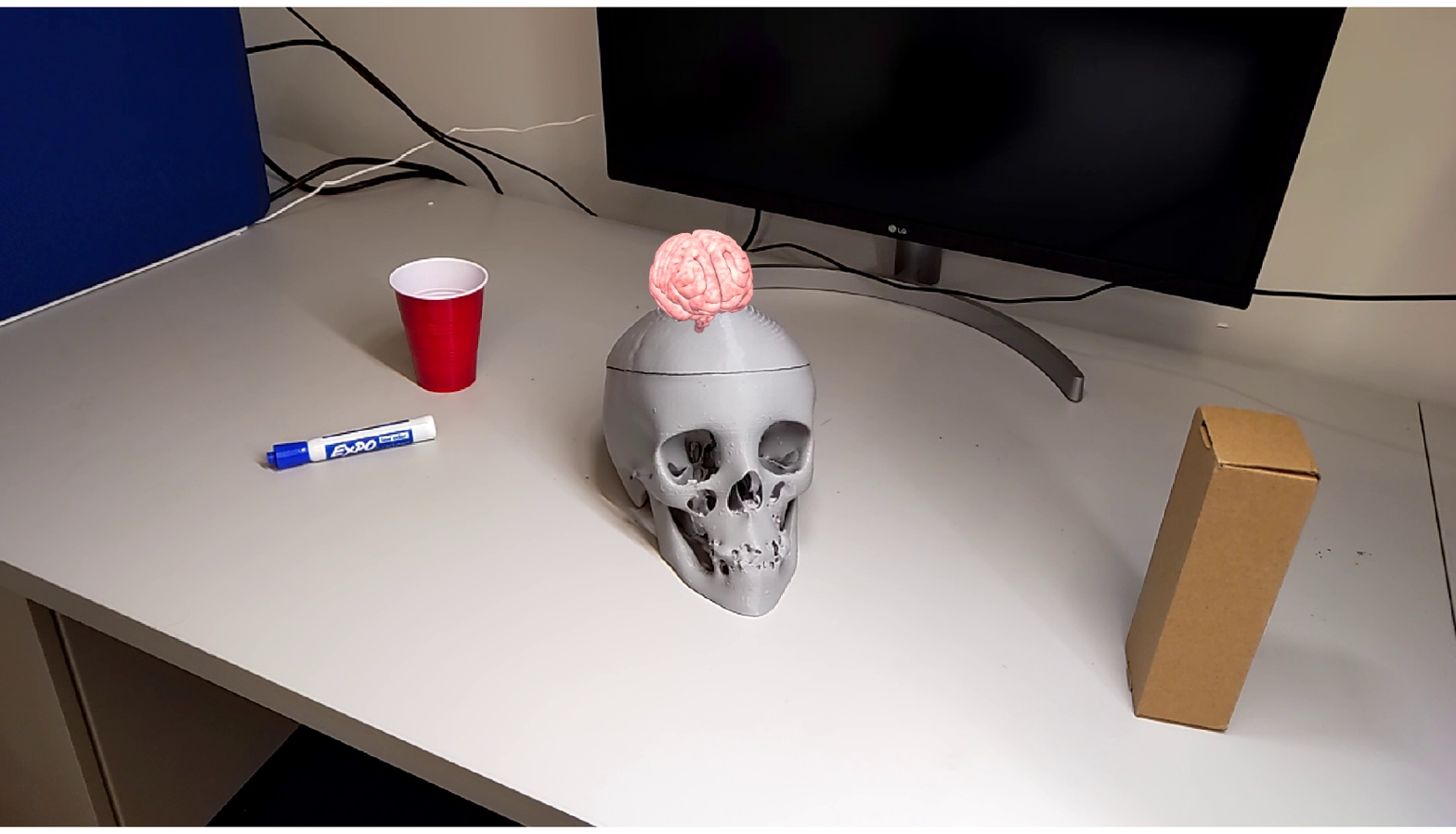}
    \label{fig:fewshot_size_1}
    \Description{Few-shot prompting instance with an assigned size appropriateness score of 1.}
\end{figure}

\noindent
\begin{center}
    \begin{minipage}{0.9\linewidth} 
\begin{shaded}
\noindent\emph{Output: ``The augmented reality content is a brain model. Score: 1''}
\newline
\noindent\emph{Input:}
\end{shaded}
\end{minipage}
\end{center}

\begin{figure}[H]
    \centering
    \includegraphics[width=0.35\linewidth]{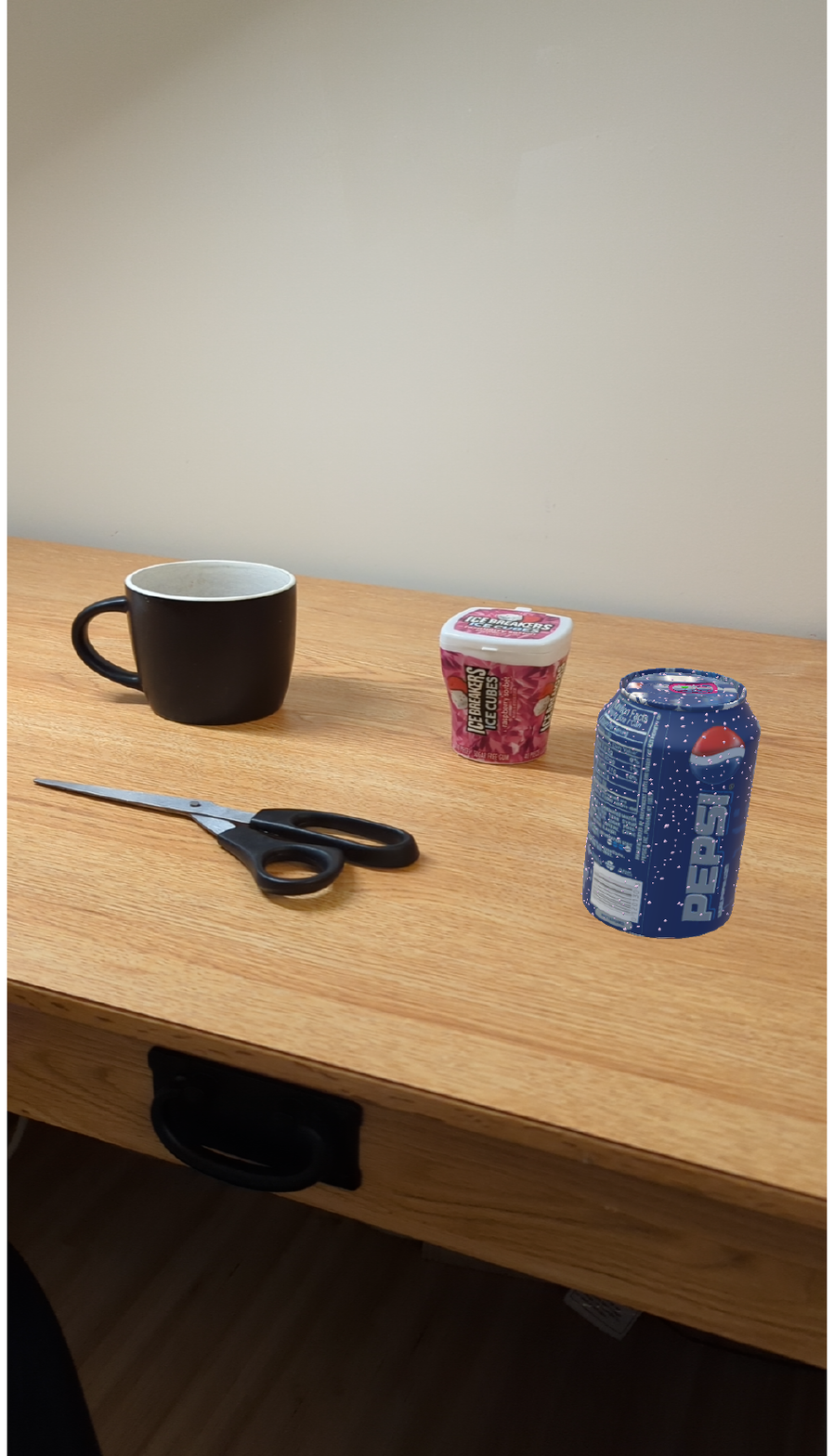}
    \label{fig:fewshot_size_5}
    \Description{Few-shot prompting instance with an assigned size appropriateness score of 5.}
\end{figure}

\noindent
\begin{center}
    \begin{minipage}{0.9\linewidth} 
\begin{shaded}
\noindent\emph{Output: ``The augmented reality content is a Pepsi can. Score: 5''}
\newline
\noindent\emph{Input:}
\end{shaded}
\end{minipage}
\end{center}

\begin{figure}[H]
    \centering
    \includegraphics[width=0.35\linewidth]{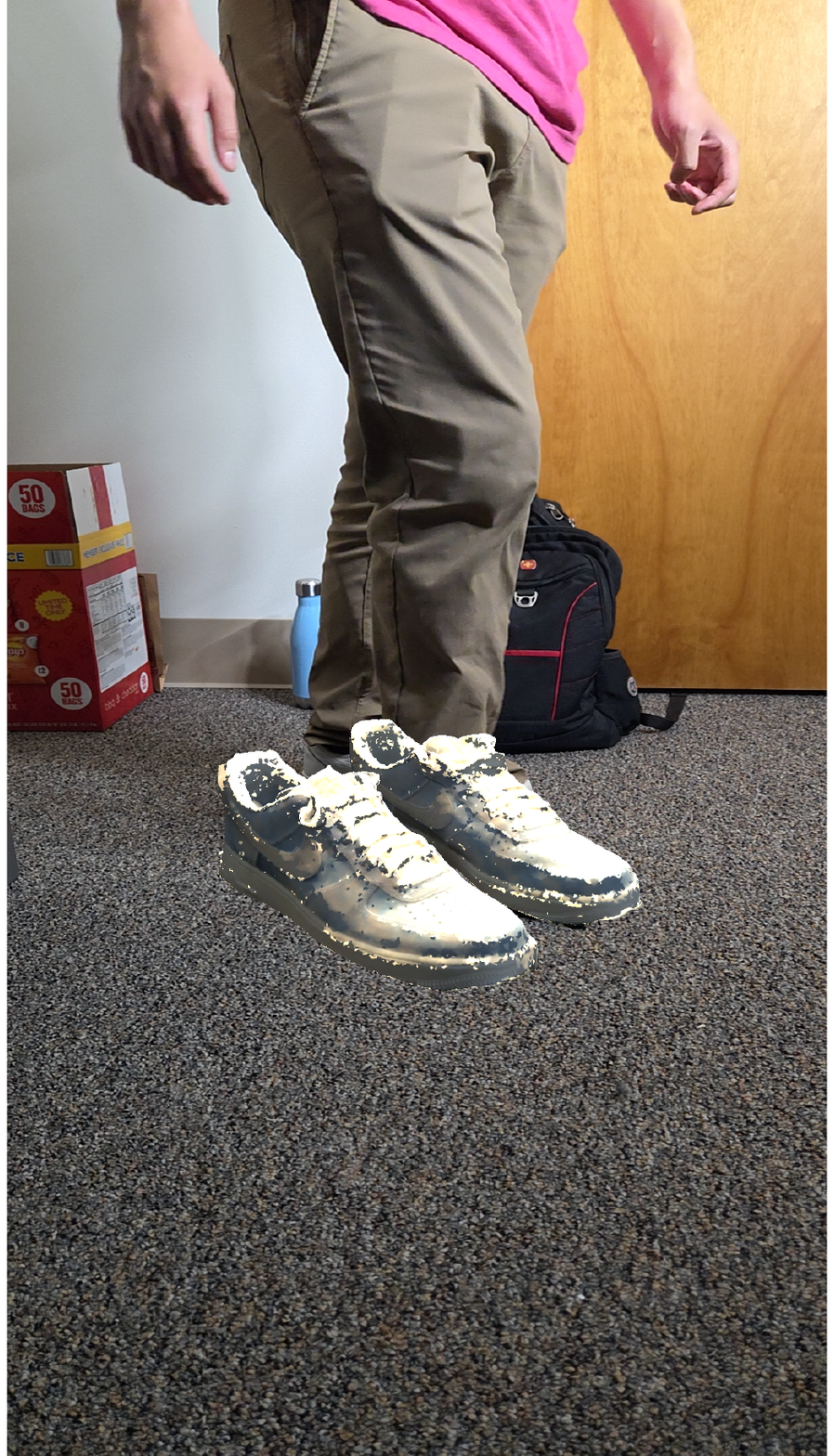}
    \label{fig:fewshot_size_3}
    \Description{Few-shot prompting instance with an assigned size appropriateness score of 3.}
\end{figure}

\noindent
\begin{center}
    \begin{minipage}{0.9\linewidth} 
\begin{shaded}
\noindent\emph{Output: ``The augmented reality content is a pair of shoes. Score: 3''}
\newline
\noindent\emph{Input: [Target image to analyze]}
\end{shaded}
\end{minipage}
\end{center}

\paragraph{Shadow Realism}
\leavevmode \\ 
\noindent
\begin{center}
    \begin{minipage}{0.9\linewidth} 
\begin{shaded}
\noindent Task: \emph{``You are an expert in observing, describing, and estimating augmented reality images. You 
are tasked with rating the shadow realism of augmented reality (AR) content. Please assess whether the virtual object’s shadow is appropriate for the AR scene. If not, explain whether the inconsistency is appropriate given the surrounding context. Rate shadow appropriateness on a 1–5 scale, where 1 means an object that should have a proper shadow instead has a clearly missing, misaligned, or inconsistent one, and 5 means the object has a relatively consistent shadow or, if intentionally non-standard, its absence is reasonable. Use cues such as whether the virtual shadow relatively aligns with nearby reference objects in direction, softness, and intensity, and whether the object is intended for conceptual visualization. Please end with a final sentence in this exact format: Score: [your rating]. You are provided with three AR scene examples and their perceived shadow realism scores.''}
\newline
\noindent\emph{Input:}
\end{shaded}
\end{minipage}
\end{center}
\begin{figure}[H]
    \centering
    \includegraphics[width=0.35\linewidth]{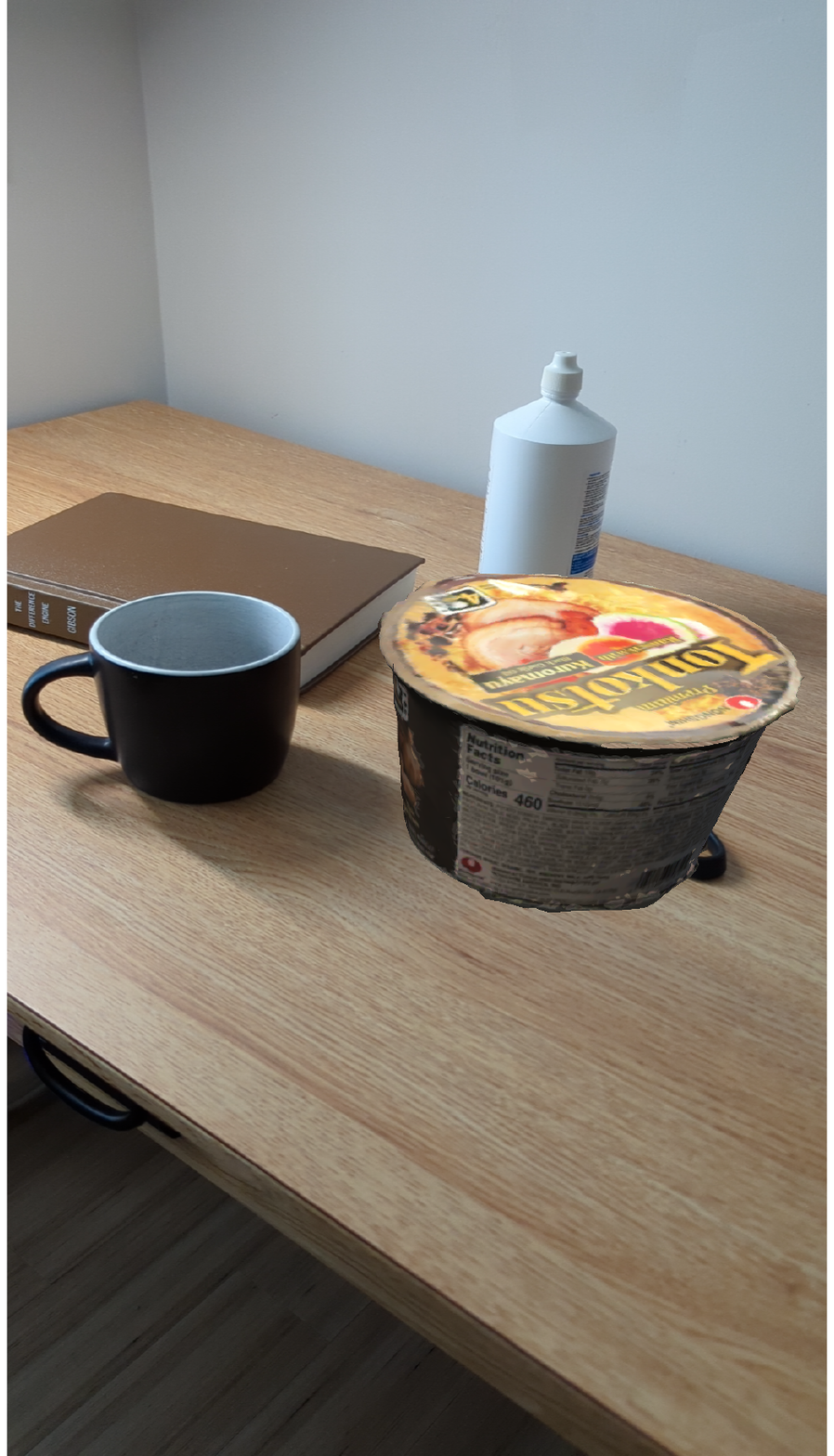}
    \label{fig:fewshot_shadow_1}
    \Description{Few-shot prompting instance with an assigned shadow realism score of 1.}
\end{figure}

\noindent
\begin{center}
    \begin{minipage}{0.9\linewidth} 
\begin{shaded}
\noindent\emph{Output: ``The augmented reality content is a noodle bowl. Score: 1''}
\newline
\noindent\emph{Input:}
\end{shaded}
\end{minipage}
\end{center}

\begin{figure}[H]
    \centering
    \includegraphics[width=0.35\linewidth]{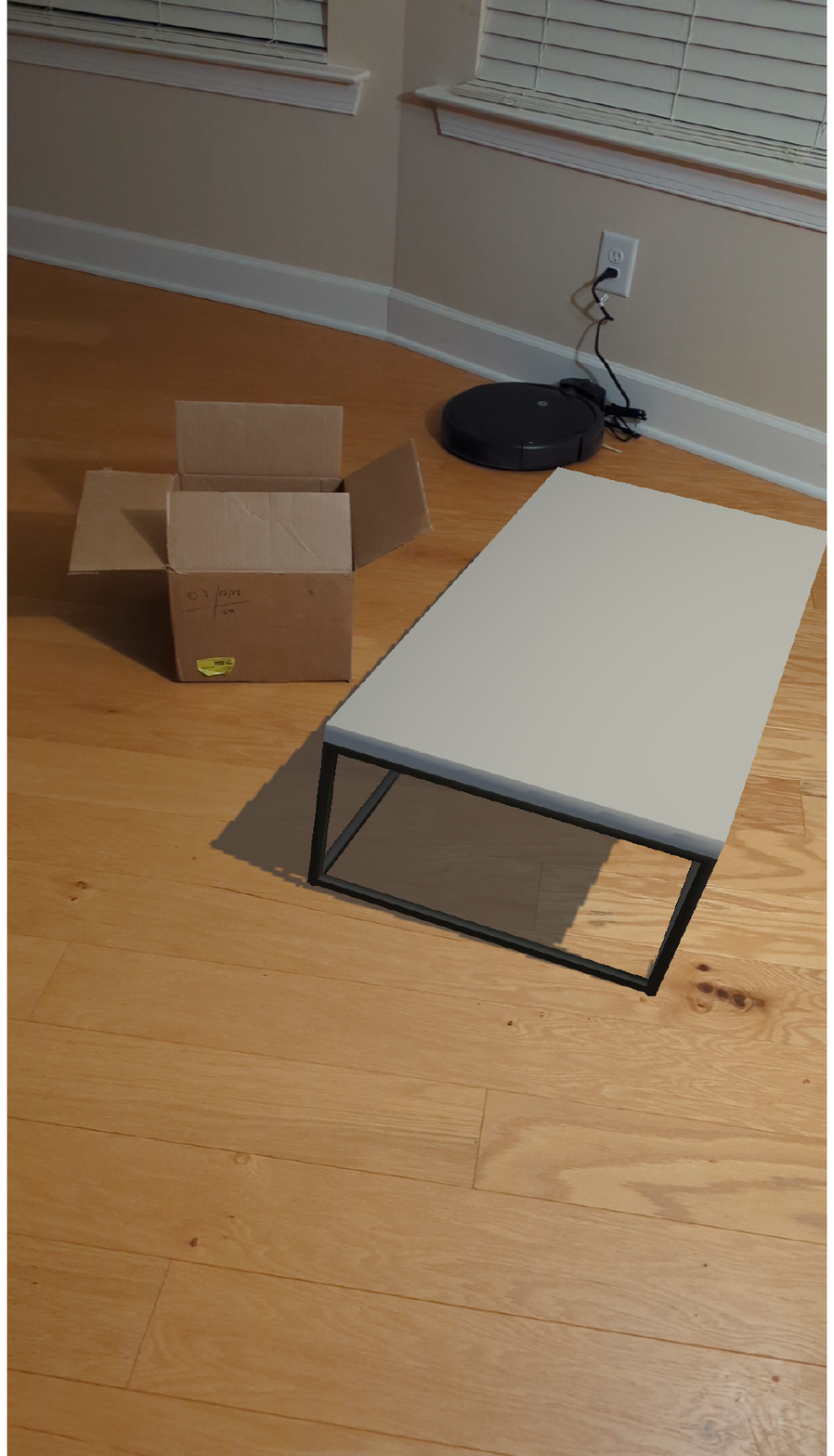}
    \label{fig:fewshot_shadow_5}
    \Description{Few-shot prompting instance with an assigned shadow realism score of 5.}
\end{figure}

\noindent
\begin{center}
    \begin{minipage}{0.9\linewidth} 
\begin{shaded}
\noindent\emph{Output: ``The augmented reality content is a coffee table. Score: 5''}
\newline
\noindent\emph{Input:}
\end{shaded}
\end{minipage}
\end{center}
\begin{figure}[H]
    \centering
    \includegraphics[width=0.35\linewidth]{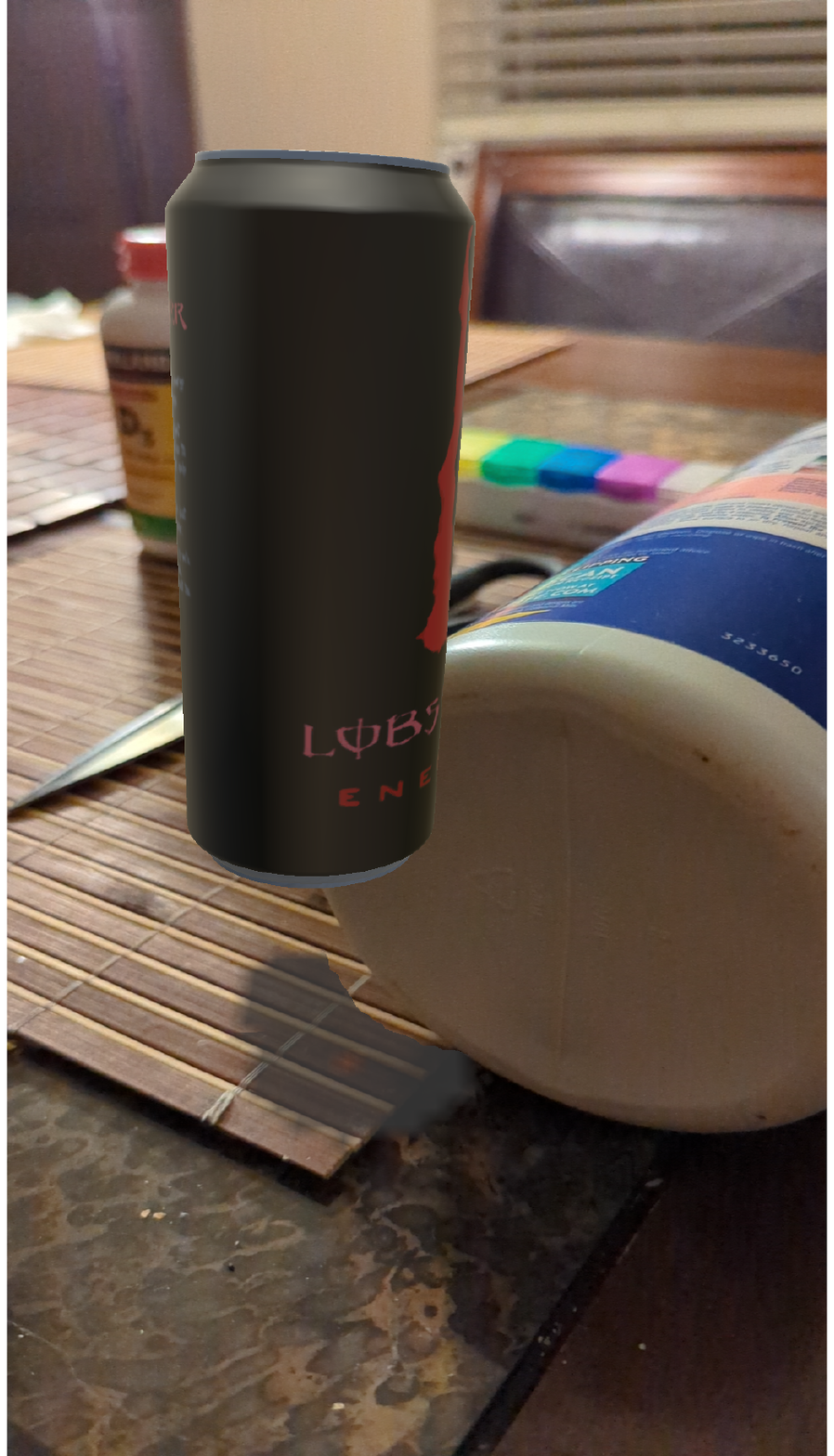}
    \label{fig:fewshot_shadow_3}
    \Description{Few-shot prompting instance with an assigned shadow realism score of 3.}
\end{figure}

\noindent
\begin{center}
    \begin{minipage}{0.9\linewidth} 
\begin{shaded}
\noindent\emph{Output: ``The augmented reality content is a beverage can. Score: 3''}
\newline
\noindent\emph{Input: [Target image to analyze]}
\end{shaded}
\end{minipage}
\end{center}

\subsection{Extended Prompting Strategy Analysis}

As stated in \Cref{sec:tables_CI_and_pvalues}, reported correlation coefficient 95\% confidence intervals and statistical significance p-values are obtained using \texttt{Pingouin}~\cite{vallat2018pingouin}. \Cref{tab:ablation_CI} and \Cref{tab:ablation_pvals} present the confidence intervals and p-values corresponding to \Cref{tab:ablation}, respectively. P-Values are adjusted, column-wise, using the Bonferroni correction method with $m_{abl} = 9$ tests, such that $p_{adj}=\min(1,p_{raw} * m_{abl})$.

\begin{table*}[htbp]
\centering
\caption{Correlation metrics between human scores and \emph{GPT-family} flagship model predictions on RateAR images, corresponding to the coefficients in~\Cref {tab:ablation}. Confidence interval format: {[lower\_bound, upper\_bound]}.}
\label{tab:ablation_CI}
\resizebox{\textwidth}{!}{%
\begin{tabular}{@{}ll|rrr|rrr|rrr}
\toprule
                             \multirow{2}{*}{\begin{tabular}[c]{@{}l@{}}\textbf{Prompt}\\ \textbf{Strategy}\end{tabular}} & \multirow{2}{*}{\textbf{VLM}} & \multicolumn{3}{c|}{\begin{tabular}[c]{@{}c@{}}\textbf{Placement}\end{tabular}}                                    & \multicolumn{3}{c|}{\begin{tabular}[c]{@{}c@{}}\textbf{Size}\end{tabular}}                                      & \multicolumn{3}{c}{\begin{tabular}[c]{@{}c@{}}\textbf{Shadow}\end{tabular}}                                             \\ \cline{3-11} 
 &     & \multicolumn{1}{c}{SRCC}               & \multicolumn{1}{c}{PLCC}               & \multicolumn{1}{c|}{KRCC}              & \multicolumn{1}{c}{SRCC}               & \multicolumn{1}{c}{PLCC}               & \multicolumn{1}{c|}{KRCC}              & \multicolumn{1}{c}{SRCC}               & \multicolumn{1}{c}{PLCC}               & \multicolumn{1}{c}{KRCC}               \\ \hline
 & GPT-o3                   & 
{[0.64, 0.77]}                   & 
{[0.63, 0.77]}                   & 
{[0.48, 0.66]}                   & 
{[0.54, 0.70]}                   & 
{[0.57, 0.72]}                   & 
{[0.40, 0.60]}                  & 
{[0.33, 0.54]}                   & 
{[0.30, 0.52]}                   & 
{[0.23, 0.47]}                   \\
Primitive & GPT-5            & 
{[0.67, 0.79]}                   & 
{[0.66, 0.78]}                   & 
{[0.52, 0.68]}                   & 
{[0.39, 0.59]}                   & 
{[0.44, 0.62]}                   & 
{[0.26, 0.49]}                  & 
{[0.36, 0.57]}                   & 
{[0.35, 0.56]}                   & 
{[0.25, 0.48]}                   \\
& GPT-5.4            & 
{[0.57, 0.72]}                   & 
{[0.55, 0.71]}                   & 
{[0.43, 0.62]}                   & 
{[-0.10, 0.16]}                   & 
{[-0.03, 0.23]}                   & 
{[-0.11, 0.15]}                  & 
{[0.53, 0.69]}                   & 
{[0.51, 0.68]}                   & 
{[0.39, 0.59]}                   \\ \hline
 & GPT-o3            & 
{[0.68, 0.80]}                   & 
{[0.64, 0.77]}                   & 
{[0.52, 0.68]}                   & 
{[0.67, 0.79]}                   & 
{[0.69, 0.81]}                   & 
{[0.49, 0.66]}                  & 
{[0.49, 0.67]}                   & 
{[0.44, 0.63]}                   & 
{[0.36, 0.56]}                   \\
Contextual & GPT-5            & 
{[0.79, 0.87]}                   & 
{[0.74, 0.84]}                   & 
{[0.61, 0.75]}                   & 
{[0.39, 0.59]}                   & 
{[0.49, 0.66]}                   & 
{[0.27, 0.49]}                  & 
{[0.40, 0.60]}                   & 
{[0.39, 0.59]}                   & 
{[0.28, 0.50]}                   \\
& GPT-5.4            & 
{[0.65, 0.78]}                   & 
{[0.62, 0.76]}                   & 
{[0.49, 0.67]}                   & 
{[0.08, 0.33]}                   & 
{[0.22, 0.46]}                   & 
{[0.03, 0.28]}                  & 
{[0.52, 0.68]}                   & 
{[0.49, 0.66]}                   & 
{[0.38, 0.58]}                   \\ \hline
 & GPT-o3            & 
{[0.65, 0.78]}                   & 
{[0.65, 0.78]}                   & 
{[0.49, 0.66]}                   & 
{[0.57, 0.73]}                   & 
{[0.65, 0.78]}                   & 
{[0.41, 0.61]}                  & 
{[0.44, 0.63]}                   & 
{[0.39, 0.59]}                   & 
{[0.31, 0.53]}                   \\
Few-shot & GPT-5            & 
{[0.75, 0.84]}                   & 
{[0.77, 0.86]}                   & 
{[0.57, 0.72]}                   & 
{[0.48, 0.66]}                   & 
{[0.54, 0.70]}                   & 
{[0.34, 0.55]}                  & 
{[0.24, 0.47]}                   & 
{[0.27, 0.49]}                   & 
{[0.15, 0.39]}                   \\
& GPT-5.4            & 
{[0.61, 0.75]}                   & 
{[0.60, 0.74]}                   & 
{[0.45, 0.64]}                   & 
{[0.22, 0.45]}                   & 
{[0.32, 0.54]}                   & 
{[0.12, 0.37]}                  & 
{[0.54, 0.70]}                   & 
{[0.50, 0.67]}                   & 
{[0.38, 0.58]}                   \\ \bottomrule
\end{tabular}}
\end{table*}

\begin{table*}[htbp]
\centering
\caption{Correlation coefficient statistical significance p-values between human scores and \emph{GPT-family} flagship model predictions on RateAR images, corresponding to the coefficients in~\Cref {tab:ablation}. P-Value format: {$p_{adj} \ (p_{raw})$; for highly significant results (i.e., $p_{adj}<.001$), only the adjusted p-value is reported.}}
\label{tab:ablation_pvals}
\begin{tabular}{@{}ll|ccc|ccc|ccc}
\toprule
                             \multirow{2}{*}{\begin{tabular}[c]{@{}l@{}}\textbf{Prompt}\\ \textbf{Strategy}\end{tabular}} & \multirow{2}{*}{\textbf{VLM}} & \multicolumn{3}{c|}{\begin{tabular}[c]{@{}c@{}}\textbf{Placement}\end{tabular}}                                    & \multicolumn{3}{c|}{\begin{tabular}[c]{@{}c@{}}\textbf{Size}\end{tabular}}                                      & \multicolumn{3}{c}{\begin{tabular}[c]{@{}c@{}}\textbf{Shadow}\end{tabular}}                                             \\ \cline{3-11} 
 &     & \multicolumn{1}{c}{SRCC}               & \multicolumn{1}{c}{PLCC}               & \multicolumn{1}{c|}{KRCC}              & \multicolumn{1}{c}{SRCC}               & \multicolumn{1}{c}{PLCC}               & \multicolumn{1}{c|}{KRCC}              & \multicolumn{1}{c}{SRCC}               & \multicolumn{1}{c}{PLCC}               & \multicolumn{1}{c}{KRCC}               \\ \hline
 & GPT-o3                   & 
{<.001}                   & 
{<.001}                   &
{<.001}                   &
{<.001}                   &
{<.001}                   &
{<.001}                   &
{<.001}                   &
{<.001}                   &
{<.001}                   \\
Primitive & GPT-5            & 
{<.001}                   & 
{<.001}                   &
{<.001}                   &
{<.001}                   &
{<.001}                   &
{<.001}                   &
{<.001}                   &
{<.001}                   &
{<.001}                   \\
& GPT-5.4            & 
{<.001}                   & 
{<.001}                   &
{<.001}                   & 
{>.999 (.666)}                   & 
{>.999 (.135)}                   & 
{>.999 (.677)}                  & 
{<.001}                   &
{<.001}                   &
{<.001}                   \\ \hline
 & GPT-o3            & 
{<.001}                   & 
{<.001}                   &
{<.001}                   &
{<.001}                   &
{<.001}                   &
{<.001}                   &
{<.001}                   &
{<.001}                   &
{<.001}                   \\
Contextual & GPT-5            & 
{<.001}                   & 
{<.001}                   &
{<.001}                   &
{<.001}                   &
{<.001}                   &
{<.001}                   &
{<.001}                   &
{<.001}                   &
{<.001}                   \\
& GPT-5.4            & 
{<.001}                   & 
{<.001}                   &
{<.001}                   & 
{.014 (.002)}                   & 
{<.001}                   & 
{.022 (.002)}                  & 
{<.001}                   &
{<.001}                   &
{<.001}                   \\ \hline
 & GPT-o3            & 
{<.001}                   & 
{<.001}                   &
{<.001}                   &
{<.001}                   &
{<.001}                   &
{<.001}                   &
{<.001}                   &
{<.001}                   &
{<.001}                   \\
Few-shot & GPT-5            & 
{<.001}                   & 
{<.001}                   &
{<.001}                   &
{<.001}                   &
{<.001}                   &
{<.001}                   &
{<.001}                   &
{<.001}                   &
{<.001}                   \\
& GPT-5.4            & 
{<.001}                   & 
{<.001}                   &
{<.001}                   &
{<.001}                   &
{<.001}                   &
{<.001}                   &
{<.001}                   &
{<.001}                   &
{<.001}                   \\ \bottomrule
\end{tabular}
\end{table*}

\end{document}